\documentclass{article}

\usepackage[preprint, nonatbib]{neurips_2026}

\usepackage[utf8]{inputenc}
\usepackage[T1]{fontenc}
\usepackage{hyperref}
\usepackage{url}
\usepackage{booktabs}
\usepackage{amsmath}
\usepackage{amssymb}
\usepackage{amsfonts}
\usepackage{nicefrac}
\usepackage{microtype}
\usepackage{xcolor}
\usepackage{graphicx}
\usepackage{xspace}
\usepackage{enumitem}

\graphicspath{{./}{../figs/}{figs/}}

\newcommand{\eg}{e.g.,\xspace}

\title{Encoder Winners Do Not Reliably Transfer Across VLA Backbone
       Scale:\\ A Frozen-Backbone Grafting Diagnostic}

\author{%
  Qingping Zeng \\
  Tsinghua University \\
  \texttt{bbai8083@gmail.com}
  \And
  Fei She \\
  Tsinghua University \\
}

\begin{document}

\maketitle

\begin{abstract}
Vision-language-action (VLA) policies typically inherit their vision
encoder from upstream VLM releases, but it is unclear whether an
encoder choice validated on a small VLA transfers to a larger
backbone. We introduce a \emph{frozen-backbone grafting} diagnostic:
the vision tower of a released VLA is replaced by a candidate encoder
under a fixed protocol (adaptive average pooling, LayerNorm, and a
single trainable linear projector), with the language model and
action expert frozen. Across four encoders, two LIBERO suites, two
backbones (SmolVLA-450M and $\pi_{0.5}$-3.3B), and two-to-three seeds
per cell ($40$ main grafting runs plus native, LoRA, pooling, and
zero-/shuffled-image controls, all scored by offline action MSE),
the small-backbone winner \emph{does not reliably select}
the large-backbone top tier: SigLIP is best on SmolVLA across both
suites, while on $\pi_{0.5}$ DINOv2-small leads the spatial suite and
the object suite is a seed-sensitive near-tie band; three of the four
backbone-suite comparisons (and $11$ of $12$ seed-level cells) support
backbone-dependent rankings. The grafting wrapper is itself non-neutral with opposite
sign across backbones ($+45$--$56\%$ MSE on the SmolVLA native tower,
$-50$--$52\%$ on $\pi_{0.5}$), so all conclusions are conditional on
the fixed grafting protocol. We position frozen grafting as a cheap
target-backbone diagnostic to run before committing to an encoder at
scale, not as a closed-loop deployment claim.
\end{abstract}

\section{Introduction}
\label{sec:intro}

Vision-language-action (VLA) models have moved from sub-billion-parameter
policies such as SmolVLA~\cite{smolvla2024} to multi-billion-parameter
foundation policies such as $\pi_{0.5}$~\cite{pi05_2024} and
OpenVLA~\cite{openvla2024}. Most of the engineering effort in this scaling
trajectory has focused on the language backbone, the action expert head, and
the demonstration mixture, while the visual encoder has typically been inherited
from each release without ablation: SmolVLA ships with the SmolVLM vision tower,
$\pi_{0.5}$ ships with the PaliGemma SigLIP tower, and OpenVLA fuses
SigLIP and DINOv2 features. Whether the encoder choice validated on a
small VLA backbone remains the right choice once the language and
action stack is scaled by an order of magnitude is an open empirical
question, and the cost of answering it sloppily is considerable: a
wrong encoder pick wastes pretraining compute and pollutes
downstream finetune curves.

A growing line of work studies encoder choice at the small-VLA scale.
SmolVLA-EdgeBench and the recent VLM4VLA study~\cite{vlm4vla2024} sweep
multiple vision backbones during VLA pretraining and report
encoder-specific performance gaps; OpenVLA-OFT~\cite{openvla_oft2024}
investigates how finetuning recipes interact with the fused
SigLIP+DINOv2 tower; Theia~\cite{theia2024} distills eight robot-relevant
encoders into a single backbone. None of these works isolate
the \emph{backbone scale} variable: the encoder is always co-trained with
the rest of the policy, and any ranking observed therefore conflates
encoder quality with backbone-encoder co-adaptation. As a result, when a
practitioner picks up $\pi_{0.5}$ or another large released VLA, there
is no controlled evidence telling her whether the small-VLA top-1
encoder will hold or flip.

In this work we run a deliberately narrow controlled study to isolate
that single question. We introduce a \emph{frozen-backbone grafting}
protocol: given a pretrained VLA, we (i) freeze the language model, the
action expert, and the original vision tower, (ii) attach a candidate
encoder via a deterministic adaptive-avg-pool plus linear projector, and
(iii) train only the projector ($0.37$M--$1.58$M parameters depending
on the encoder output dimension and the target backbone hidden size)
for $2{,}000$ steps with batch size $8$ (SmolVLA) or effective batch
size $8$ ($\pi_{0.5}$, micro-batch $2$ with gradient accumulation $4$)
at learning rate $10^{-4}$. We graft four representative encoders
(\texttt{siglip\_base}, \texttt{dinov2\_small}, \texttt{fastvit\_sa12},
\texttt{repvit\_m1}) onto two released backbones (SmolVLA-450M and
$\pi_{0.5}$-3.3B) across two LIBERO image task suites
(\texttt{libero\_spatial}, \texttt{libero\_object}), with two-to-three
seeds per cell (three seeds for the SigLIP and DINOv2 cells, two seeds
for the FastViT and RepViT cells), evaluated by offline action
mean-squared-error on an episode-split validation set ($24{,}913$
training windows / $6{,}457$ validation windows per suite). Because
the SmolVLA SO-100 checkpoint and the LIBERO Franka embodiment
disagree on the action space, native closed-loop success collapses to
zero for both backbones; we therefore treat this study as a controlled
offline diagnostic rather than a closed-loop benchmark, and we report
exclusively offline action MSE.

The 40-configuration grid yields three diagnostic observations.
First, the per-encoder top-1 choice is \emph{backbone-dependent and
suite-dependent}: on SmolVLA the semantic SigLIP encoder attains the
lowest mean MSE on both suites (spatial mean MSE $0.0706$, object
mean MSE $0.0628$, ahead of DINOv2 at $0.0734$ / $0.0675$); on
$\pi_{0.5}$-\texttt{spatial} the geometry-oriented DINOv2 encoder
attains the lowest mean MSE ($0.0256$, ahead of SigLIP at $0.0267$
and FastViT at $0.0283$); on $\pi_{0.5}$-\texttt{object} the three
strongest encoders sit in a near-tie band (SigLIP $0.02149$, DINOv2
$0.02166$, FastViT $0.02206$, all within $2.7\%$ relative). Across the
four backbone-suite SigLIP-vs-DINOv2 comparisons, three support the
backbone-dependent top-tier pattern; the lone exception is
$\pi_{0.5}$-\texttt{libero\_object}, where the
direction is supported at $2/3$ seeds with seed-44 reversing. At the
finer cell level the expected direction holds in $11/12$ pairwise
(backbone, suite, seed) cells, which we report descriptively given
shared seeds across suites; the single cell-level reversal is at
$\pi_{0.5}$-\texttt{object}-seed44 (SigLIP $0.02018$ below DINOv2
$0.02288$). On $\pi_{0.5}$-\texttt{libero\_spatial}-seed44 the
DINOv2-vs-SigLIP gap is a numerical near-tie ($0.02510$ DINOv2 vs
$0.02511$ SigLIP, $\Delta = 0.00001$ MSE, $0.04\%$ relative), and we
flag this comparison as near-tie rather than as a substantive
DINOv2 win, since the gap is at the edge of cross-seed noise. The
cross-backbone
Spearman rank correlation remains positive because the bottom of the
pool transfers, but the small-backbone top-1 choice does not reliably
select the large-backbone top tier, and on $\pi_{0.5}$-\texttt{object}
it is not even stable to a single additional seed on the large
backbone.
Second, the grafting harness itself --- frozen backbone, deterministic
pool, single-layer projector --- is a reproducible diagnostic that
costs under six GPU-hours per cell on a single GB10 node (minutes for
SmolVLA, about five hours for $\pi_{0.5}$) and can
be re-run as new VLA backbones are released. Third, the offline
action-MSE diagnostic, while weaker than closed-loop success, gives a
usable signal in exactly the regime where closed-loop rollouts are
blocked by embodiment mismatch between the pretrained VLA checkpoint
and the available simulator, which is increasingly common as VLAs
ship without matched simulators.

Our contributions are:
\begin{itemize}
\item \textbf{Small-backbone encoder winners do not reliably select
the large-backbone top tier.}
Under a controlled frozen-backbone grafting protocol, the lowest-MSE
encoder on SmolVLA-450M (SigLIP) is \emph{not} the lowest-MSE encoder
on $\pi_{0.5}$-3.3B's \texttt{libero\_spatial} suite (DINOv2-small),
and on $\pi_{0.5}$-\texttt{libero\_object} the top three encoders sit
in a near-tie band where the top-1 identity is not stable under seed
perturbation. Across the four backbone-suite SigLIP-vs-DINOv2
comparisons, three support the backbone-dependent top-tier
pattern --- our load-bearing inferential summary --- and the finer
$11/12$ seed-level cell breakdown is reported only as descriptive support
(\autoref{tab:directional}) and not as independent inferential evidence
given shared seeds and projector initialization. We do not interpret this as a full rank inversion ---
the Spearman correlation stays positive because the worst encoder
stays worst --- only as the observation that the small-backbone
encoder winner does not reliably select the large-backbone top tier
under the offline diagnostic.
\item \textbf{Open grafting harness.} We release an encoder-swap
wrapper for SmolVLA and $\pi_{0.5}$ that hooks the vision pathway at a
single point, normalises feature shape via a deterministic
adaptive-avg-pool to a fixed token count, and trains only a linear
projector, so that new encoders or new backbones can be plugged in
without touching the language or action stack.
\item \textbf{Offline diagnostic under embodiment mismatch.} We adopt
offline action-MSE on episode-split held-out windows as the
primary metric, which allows controlled comparisons even when
closed-loop rollout success is collapsed by embodiment mismatch
between the released VLA checkpoint and the target simulator. We do
not claim closed-loop success improvement.
\end{itemize}

We position this work narrowly: we do \emph{not} claim sim-to-real
transfer, we do \emph{not} train a selector network, and we do
\emph{not} attempt encoder distillation. While VLM4VLA~\cite{vlm4vla2024}
studies the choice of VLM at \emph{training time}, we study a post-hoc
encoder swap under a frozen language model and frozen action expert,
which we view as the operationally relevant question for practitioners
who inherit an existing VLA checkpoint. We see our results as a
cautionary diagnostic: an encoder ranking obtained on one VLA backbone
is not, on its own, sufficient evidence for the same top-1 choice on a
different released backbone, and a small grafting sweep on the target
backbone is a cheap way to confirm or refute that transfer.

\section{Related Work}
\label{sec:related}

\paragraph{Vision encoder choice during VLA pretraining.}
The closest prior work to ours is VLM4VLA~\cite{vlm4vla2024}, which
systematically varies the underlying vision-language model when training a
VLA from scratch and reports task-conditional rankings between several
encoder--LM pairings. The setting differs from ours in two important
ways. First, VLM4VLA varies the entire VLM jointly, so the language
backbone and the visual encoder are co-trained and their contributions
to the final ranking are entangled. Second, VLM4VLA reports a
training-time choice, whereas we report a \emph{post-hoc} swap on a
released, frozen checkpoint. Practitioners who inherit a pretrained
$\pi_{0.5}$ or SmolVLA checkpoint cannot redo VLM pretraining; the
question they face is whether to swap the vision tower on the inherited
backbone, which is exactly the variable we isolate. Our grafting
protocol therefore complements VLM4VLA: we keep the language model and
the action expert frozen and only change the encoder plus a small
projector, isolating the encoder-quality axis from the LM-encoder
co-adaptation axis.

\paragraph{Finetuning recipes for fused encoders.}
OpenVLA-OFT~\cite{openvla_oft2024} studies how different finetuning
recipes (LoRA, full finetune, action-head-only) interact with the fused
SigLIP+DINOv2 tower of OpenVLA. Their focus is on adapting the
\emph{existing} encoder stack to downstream tasks rather than asking
whether the stack itself is the right one; the encoder identity is
held constant across their conditions. We hold the finetuning recipe
constant (projector-only training for 2{,}000 steps) and vary the
encoder identity instead, which is orthogonal to their axis. The
projector-only regime is also deliberately conservative: it removes
LoRA-recipe confounds when comparing across encoders, at the price of
a weaker absolute performance ceiling.

\paragraph{Zero-data sim-to-real and benchmark stress tests.}
VLA-0~\cite{vla02024} pushes zero-shot transfer of an action policy to
real hardware without any in-domain finetuning. We deliberately avoid
sim-to-real claims: our LIBERO checkpoints are not deployed on hardware
in this work, and the offline action-MSE we report is a diagnostic on
the LIBERO image distribution, not a deployment metric. We also do
\emph{not} share VLA-0's zero-data assumption; we always train the
projector for 2{,}000 steps before evaluating an encoder, since the
question we ask is which encoder is the better partner under a
controlled finetune budget, not which encoder is the best frozen
feature extractor. On the evaluation side, LIBERO-plus and
LIBERO-PRO~\cite{liberoplus2024,liberopro2024} introduce perturbed
versions of the LIBERO image suites (lighting, texture, camera) to
stress-test policies under distribution shift. Our current grid uses
clean LIBERO and reports MSE only on the clean held-out split; we view
LIBERO-plus as the natural next axis for the same grafting harness
and leave it to a follow-up.

\paragraph{Multi-encoder distillation.}
Theia~\cite{theia2024} distills eight robot-relevant encoders into a
single student encoder and reports gains over the strongest teacher.
The Theia agenda is to \emph{compress} the encoder choice problem into
a single trained backbone. Ours is the opposite: we keep encoders
separate and ask whether the per-encoder ranking is stable across
VLA backbones, which is exactly the diagnostic that motivates a
distillation effort in the first place. The two directions are
compatible. If our top-1-non-reliable-transfer observation holds at
additional released backbones, a distillation step like Theia's would
need to be re-run per backbone rather than re-used across backbones;
if the top-1 choice turns out to be stable beyond the two backbones we
test, a single distilled encoder becomes more attractive.

\paragraph{Foundation VLA models we graft onto.}
We graft onto SmolVLA~\cite{smolvla2024}, $\pi_{0.5}$~\cite{pi05_2024},
and we situate our results relative to OpenVLA~\cite{openvla2024}.
SmolVLA pairs a 450M SmolVLM backbone with a small action expert and
is intentionally edge-oriented; the released checkpoint we use is
trained on SO-100 manipulation data. $\pi_{0.5}$ pairs a PaliGemma-3B
vision-language backbone with a flow-matching action expert at roughly
3.3B total parameters and is the largest backbone in our grid.
OpenVLA fuses SigLIP and DINOv2 features for a 7B-class policy; we do
not graft onto OpenVLA in the current grid because its native
embodiment is closer to ours than $\pi_{0.5}$ is, but the harness is
designed to extend to it. Across all three families, the released
checkpoints inherit the vision encoder choice from upstream VLM
pretraining without published ablations of alternative encoders on
the released backbone, which is the gap our protocol is built to
probe.

\paragraph{Lightweight vision backbones in robotics.}
The four encoders in our grid span the design axes that practitioners
care about. SigLIP is the canonical semantic ViT used by PaliGemma and
inherited by $\pi_{0.5}$; DINOv2 is the canonical geometric ViT and
appears in OpenVLA's fused tower; FastViT-SA12 represents
Apple-style attention-convolution hybrids tuned for on-device latency;
RepViT-M1 represents structural reparameterisation tuned for
embedded CNN deployment. All four are sub-100M-parameter backbones
evaluated under a unified $224{\times}224$ input path, which makes
them realistic candidates for a
robot operator who needs to swap encoders without growing the
deployment envelope. Larger encoders such as DINOv2-Giant or
SigLIP-Large would shift the latency-quality Pareto and are an
obvious next axis for the harness; we exclude them from the current
grid to keep the cross-backbone comparison clean and to keep the
per-cell compute budget under six GPU-hours.

\section{Method}
\label{sec:method}

\begin{figure}[t]
  \centering
  \includegraphics[width=0.95\columnwidth]{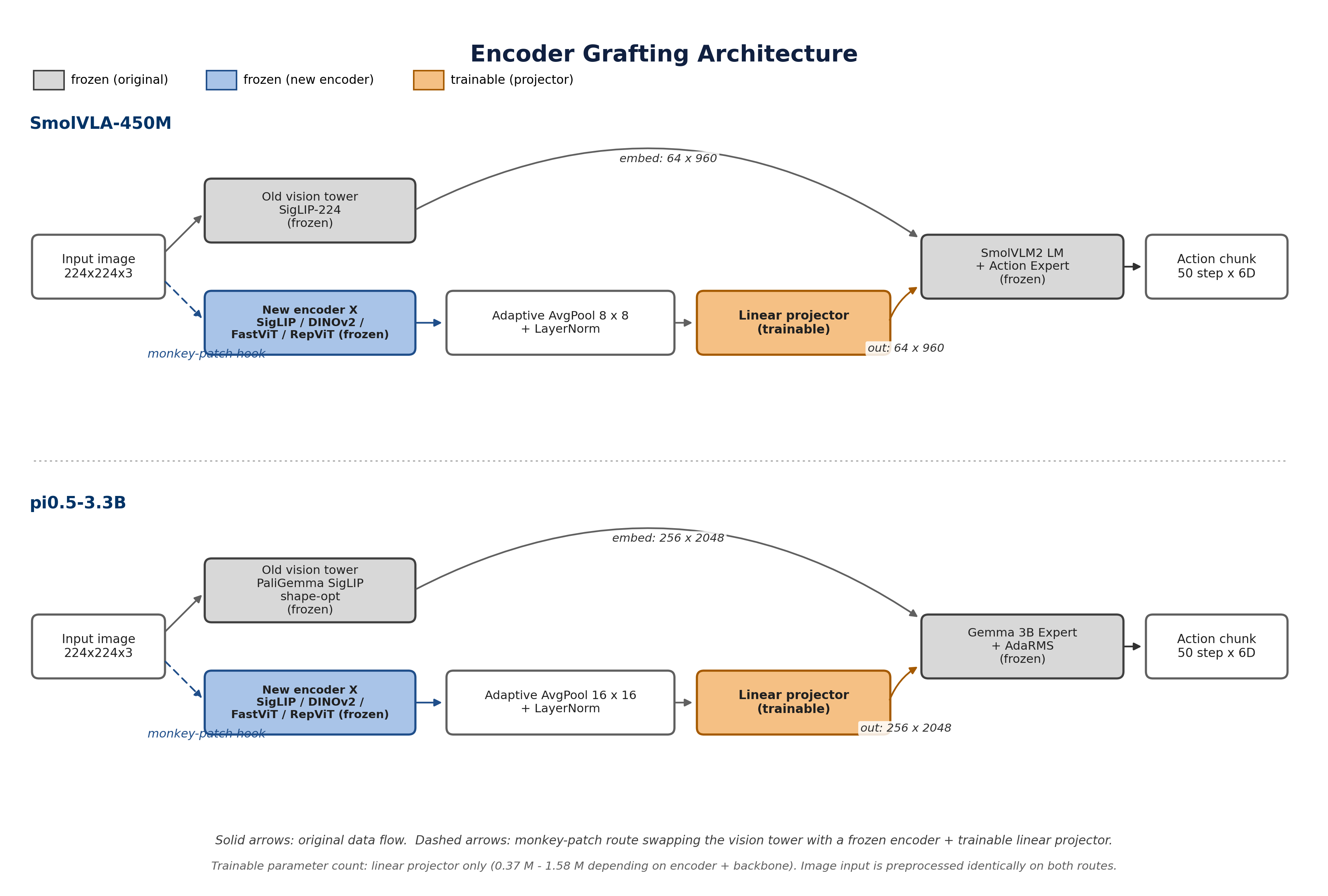}
  \caption{Frozen-backbone grafting pipeline. The language model, action
  expert, and original vision tower of each VLA are kept frozen; the
  alternative vision encoder is connected through a lightweight projector
  ($0.37$M--$1.58$M parameters depending on the encoder output dimension
  and the target backbone hidden size) trained for $2{,}000$ steps.}
  \label{fig:method_arch}
\end{figure}

We compare vision encoders inside two vision-language-action (VLA) policies by
\emph{grafting} alternative encoders into otherwise unchanged backbones and
training only a thin projector. This design isolates the contribution of the
vision encoder from that of the language model and action expert.
The overall pipeline is shown in \autoref{fig:method_arch}.

\subsection{Backbone architectures}
\label{sec:method:arch}

\paragraph{SmolVLA-450M.}
SmolVLA is built on top of SmolVLM2-500M-Video-Instruct\cite{smolvlm2024}.
Its vision-language module (\texttt{vlm\_with\_expert}) wraps a
SigLIP base/16 224\cite{siglip2023} vision tower followed by a
\textit{pixel-shuffle} connector with $4{\times}$ spatial downscale,
producing 64 visual tokens of hidden dimension 960.
A 100M-parameter action expert is co-attended with the visual tokens to
predict a chunked action sequence. We patch the visual entry point
\texttt{policy.model.vlm\_with\_expert.embed\_image} so that the
downstream LM and action expert see exactly the same number of tokens
and the same hidden dimension as in the native model.

\paragraph{$\boldsymbol{\pi}_{0.5}$-3.3B.}
$\pi_{0.5}$\cite{pi05_2024} extends $\pi_0$ with a PaliGemma 3B backbone
\cite{paligemma2024}: a shape-optimized SigLIP vision tower feeds a
\texttt{multi\_modal\_projector} into a Gemma expert with hidden
dimension 2048. Visual features enter the expert as a sequence of 256
tokens. We replace \texttt{paligemma.model.vision\_tower.forward} with
our grafted encoder so the projector and expert receive
the same $(B,256,2048)$ tensor shape as in the original checkpoint.

\subsection{Encoder grafting}
\label{sec:method:graft}

Both backbones expose a single visual entry point that returns a sequence
of token embeddings. We replace that entry point at runtime via a
monkey-patch so the encoder swap is completely transparent to the
downstream policy code. The grafted module consists of three stages:

\begin{enumerate}
  \item \textbf{Foreign encoder forward.} We run the candidate
        vision encoder (\eg SigLIP, DINOv2, FastViT, RepViT) on the
        $224{\times}224$ RGB image and read the final feature map.
        The encoder is loaded with pretrained weights and \emph{frozen}.
  \item \textbf{Deterministic spatial pooling.} The feature map is
        reshaped to $(B, C, H, W)$ and passed through an
        \texttt{AdaptiveAvgPool2d} that targets $8{\times}8$ for SmolVLA
        (64 tokens) or $16{\times}16$ for $\pi_{0.5}$ (256 tokens).
        Avoiding cross-attention pooling keeps the protocol simple and
        deterministic; we ablate this choice in
        \autoref{sec:exp:pool}.
  \item \textbf{Token-wise LayerNorm + linear projector.} The pooled
        tokens are passed through \texttt{LayerNorm}\cite{ba2016layer}
        and a single linear layer to the backbone hidden size (960 or
        2048). This is the \emph{only} module with trainable parameters
        ($0.37$M--$1.58$M depending on encoder and backbone).
\end{enumerate}

The encoder, the language model and the action expert are all kept
\emph{frozen}. This is a deliberate choice: a fair cross-encoder
comparison requires that any observed gap be attributable to the
encoder itself rather than to gradients flowing through the language
backbone. Freezing the original vision tower is not required because we
have already removed it from the computation graph; we leave its
parameters in memory only to keep checkpoint loading paths intact.

\subsection{Training protocol}
\label{sec:method:train}

We train each grafted policy on the corresponding LeRobot
\texttt{libero\_\{spatial,object\}\_image} dataset\cite{liberobench2023},
which contains $24{,}913$ training windows and $6{,}457$ validation
windows of $T{=}50$-step action chunks, split by episode so that no
episode contributes windows to both partitions. We sample $T{=}50$-step
action chunks using delta-timestamps $\{i/10\}_{i=0}^{49}$, matching
the action head configuration of the released checkpoints.

We use AdamW\cite{loshchilov2017adamw} with learning rate $10^{-4}$,
weight decay $10^{-4}$, no learning rate schedule, $\beta=(0.9, 0.999)$,
and \textsc{bf16} mixed precision. SmolVLA runs with batch size $8$ on
a single GPU; $\pi_{0.5}$ uses batch size $2$ with $4$-step gradient
accumulation for an effective batch size of $8$. Each configuration is
trained for a fixed budget of $2{,}000$ optimizer steps; the
training loss curves that justify the choice of $2{,}000$ steps are
shown in \autoref{fig:supp_loss}, and the wording throughout is
therefore \emph{under a fixed $2{,}000$-step projector budget} rather
than ``best encoder absolutely''. We repeat each $(\text{encoder}, \text{suite})$ pair
under either three random seeds (seeds $\{42, 43, 44\}$ for the
SigLIP and DINOv2 cells) or two random seeds (seeds $\{42, 43\}$ for
the FastViT and RepViT cells) for a total of
$(2 \cdot 3 + 2 \cdot 2) \cdot 2 \cdot 2 = 40$ training runs.

\subsection{Evaluation protocol}
\label{sec:method:eval}

We report \emph{offline action prediction error} on the held-out
LIBERO validation split. For each batch we feed observations through
the grafted policy, sample chunked actions, and compute the mean
squared error and L1 error against the ground-truth chunk:
\[
  \texttt{val\_mse} \;=\;
  \frac{1}{B T D}\sum_{b,t,d}
    \bigl(\hat a_{b,t,d} - a_{b,t,d}\bigr)^{2},
\]
where $D$ is the action dimensionality. \texttt{val\_mse} is the primary
ranking metric; \texttt{val\_l1} is an auxiliary check that conclusions
are not driven by outliers. We also log per-run wall time and peak GPU
memory.

All validation runs use the same evaluation configuration across
encoders within a given backbone: identical action-sampling seed,
fixed evaluation-loader episode and frame ordering, identical
delta-timestamp window scheme, identical observation preprocessing
pipeline, and identical MSE computation. The only intentional source
of cross-run variation inside a (backbone, suite, encoder) cell is the
training seed used for projector initialization and minibatch
shuffling; cross-encoder differences within a (backbone, suite, seed)
cell therefore reflect the encoder swap rather than evaluation-side
randomization.

We deliberately do not report closed-loop success rate on LIBERO.
The released SmolVLA and $\pi_{0.5}$ checkpoints were trained on
SO-100 demonstrations; LIBERO ships with a Franka embodiment whose
action schema, gripper convention and state vector do not match.
Zero-shot closed-loop success is therefore $\approx 0$ for both native
and grafted variants. Crucially, the encoder-swap pipeline is exactly
the standard LeRobot fine-tuning path, so this mismatch \emph{does not
affect the fairness of the offline comparison}; it does mean that
absolute action errors should be interpreted within each backbone
rather than across backbones, and that no closed-loop deployment
claim is made on the basis of these MSE numbers. We discuss this
limitation and our mitigations in \autoref{sec:disc:limit}.

\subsection{Statistical protocol}
\label{sec:method:stats}

For each backbone we compute per-encoder mean \texttt{val\_mse}, and
then describe cross-backbone consistency using (i) suite-level
directional consistency of the SigLIP-vs-DINOv2 paired comparison
(the $3/4$ backbone-suite summary that we report as the
load-bearing inferential statement), supported by a descriptive
$11/12$ cell-level breakdown across the $12$ (backbone, suite, seed) cells
(\autoref{sec:exp:effect_size}), and
(ii) Spearman's $\rho$ and Kendall's $\tau$ between the per-encoder
rankings on the two backbones, reported with sign and magnitude
rather than as significance tests (\autoref{sec:exp:rank_stability}).
A \emph{top-1 non-reliable-transfer event} is declared when the top-1
encoder on the small backbone is not the top-1 encoder on the large
backbone under the same suite, \emph{or} when the seed-level top-1
identity on the large backbone is not stable to a single additional
seed. This is the load-bearing observation of this paper. Bootstrap
intervals over validation episodes are reported only to quantify
validation-set measurement noise for a fixed model, not as a substitute
for seed-level uncertainty. The encoder pool, suite list, primary
seeds ($42, 43$) and ranking metric were fixed before the $\pi_{0.5}$
matrix was run; seed $44$ was added afterwards as an additional
robustness check for the SigLIP/DINOv2 cells only. The corresponding
commit hash and freeze tag are released alongside the raw JSONL logs.

\section{Experiments}
\label{sec:experiments}

We design our study around a single falsifiable question: does the
top-1 vision encoder selected on a small released VLA backbone transfer
to a larger released VLA backbone under identical training and
evaluation protocols? This section describes the datasets, encoder
pool, baseline anchors, statistical protocol and compute used to
answer it.

\subsection{Datasets and tasks}
\label{sec:exp:data}

We use the LeRobot-formatted versions of two LIBERO suites
\cite{liberobench2023}:

\begin{itemize}
  \item \texttt{lerobot/libero\_spatial\_image}, which evaluates
        spatial reasoning over fixed object identities placed at varied
        positions and orientations.
  \item \texttt{lerobot/libero\_object\_image}, which evaluates
        object-identity robustness with fixed scene layouts.
\end{itemize}

Each suite contributes $24{,}913$ training windows and $6{,}457$
validation windows of $T{=}50$-step action chunks, split by episode so
that no episode appears in both partitions (the underlying LIBERO
suites contribute on the order of $432$ episodes each, and the
window counts above are the per-suite chunk-window totals after the
delta-timestamp sampling described in \autoref{sec:method:train}). We
treat the two suites as independent samples drawn from different task
distributions; this enables a same-backbone \emph{cross-suite}
consistency check that complements the \emph{cross-backbone} top-1
analysis.

\subsection{Vision encoders}
\label{sec:exp:enc}

We intentionally restrict the main matrix to four representative
encoders spanning semantic ViT, self-supervised geometric ViT, hybrid
edge transformer, and mobile CNN. The goal is not to exhaustively rank
all encoders, but to test whether a small-backbone top-1 choice
transfers to a larger released backbone under a controlled grafting
protocol. The four encoders span the parameter-count
$\times$ inductive-bias plane:

\begin{itemize}
  \item \textbf{SigLIP base/16 224}\cite{siglip2023} -- a strong
        semantic encoder pretrained with a sigmoid contrastive
        objective; $93$M image-tower parameters as loaded
        (\autoref{tab:confound_meta}).
  \item \textbf{DINOv2 small}\cite{oquab2023dinov2} -- a self-supervised
        ViT pretrained on a curated 142M image set, biased toward
        geometric and part-level features; $22$M parameters.
  \item \textbf{FastViT SA12}\cite{vasu2023fastvit} -- an
        edge-optimized hybrid transformer/CNN from Apple; $11$M
        parameters.
  \item \textbf{RepViT M1}\cite{wang2023repvit} -- a mobile-class
        re-parameterized convolutional network; $5$M parameters.
\end{itemize}

All encoders are loaded from \texttt{timm} with publicly available
pretrained weights and are kept frozen during training.

\subsection{Baseline anchors}
\label{sec:exp:base}

To prevent absolute-error claims from drifting from reality we run two
native-encoder anchors: the unmodified SmolVLA and unmodified
$\pi_{0.5}$ checkpoints fine-tuned on the same LIBERO data under the
same training protocol. The anchor is the upper bound on a
``no-encoder-swap'' strategy: any grafted encoder whose
\texttt{val\_mse} approaches this number is essentially recovering the
information available in the native pipeline. Because the native
vision tower is itself a SigLIP variant in both backbones, comparing
the grafted SigLIP run to the anchor isolates the cost of the
projector and pooling stages from the cost of encoder choice.

\subsection{Hardware and compute budget}
\label{sec:exp:hw}

All training runs are executed on a single NVIDIA DGX Spark GB10
workstation with $128$\,GB of unified memory, exposing one Blackwell
GPU per process. We use \textsc{bf16} mixed precision throughout.

Empirical resource usage:

\begin{itemize}
  \item SmolVLA: $\approx$\,$9$ minutes per 2{,}000-step run, peak GPU
        memory $1.85$\,GB. The full SmolVLA matrix
        ($2 \cdot 3 + 2 \cdot 2 = 10$ encoder--seed combinations
        $\times$ $2$ suites $= 20$ runs) completes in
        $\approx 3$ hours.
  \item $\pi_{0.5}$: $\approx$\,$5$ hours per 2{,}000-step run, peak
        GPU memory $14.7$\,GB. The full $\pi_{0.5}$ matrix ($20$ runs)
        requires $\approx 100$ hours of wall-clock time.
\end{itemize}

We report full ranking statistics over the complete $40$-cell matrix.
The encoder pool, suite list, seed list (\{42, 43\} fixed before the
matrix started; seed $44$ added afterwards as an additional robustness
check for the SigLIP/DINOv2 cells) and ranking metric were fixed
before the $\pi_{0.5}$ matrix began, and no encoder was added or
removed post-hoc.

\subsection{Reproducibility}
\label{sec:exp:repro}

The grafting harness, training scripts, evaluation code, raw
JSON-Lines result logs and exact dataset hashes are released alongside
this paper. Each row of the result log carries the encoder name,
backbone name, suite, seed, validation MSE, validation L1, wall time
and peak GPU memory, so all reported statistics can be re-derived from
scratch without retraining.

\subsection{Main result table}
\label{sec:exp:main_table}

\begin{figure}[t]
  \centering
  \includegraphics[width=0.95\columnwidth]{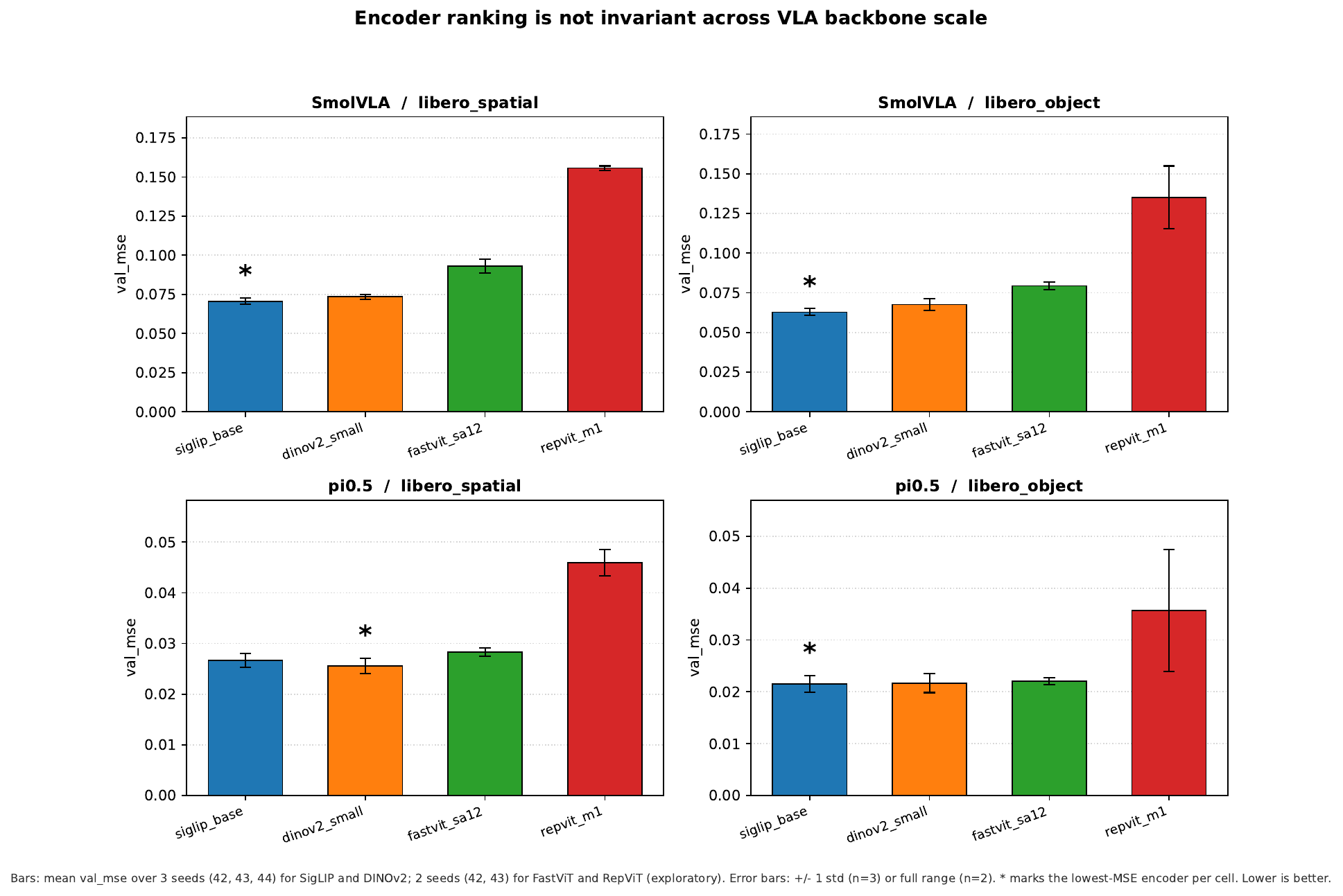}
  \caption{Per-cell offline action MSE on \texttt{libero\_spatial}
  and \texttt{libero\_object} under the frozen-backbone grafting
  protocol (the same data as \autoref{tab:main_results}). SigLIP is
  the lowest-mean-MSE encoder on both suites for
  SmolVLA-450M; DINOv2-small is the lowest-mean-MSE encoder on the
  $\pi_{0.5}$-\texttt{spatial} suite; on $\pi_{0.5}$-\texttt{object}
  the three strongest encoders sit in a near-tie band where the
  seed-44 top-1 identity reverses to SigLIP relative to the seed
  \{42, 43\} reading. SigLIP and DINOv2 are at three seeds per cell
  (seeds \{42, 43, 44\}); FastViT and RepViT are at two seeds per
  cell (seeds \{42, 43\}). Note different $y$-axis ranges across
  backbones.}
  \label{fig:ranking_matrix}
\end{figure}

\autoref{tab:main_results} reports the headline grafting matrix:
seed-averaged validation MSE with cross-seed standard deviation and
seed count $n$ for every (backbone, suite, encoder) cell, and the gap
to the per-cell best encoder, and \autoref{fig:ranking_matrix}
visualizes the same per-cell MSEs as grouped bars. Both halves of the
matrix are complete
($40$ runs total: SigLIP/DINOv2 cells use three seeds, FastViT/RepViT
cells use two seeds).
Two facts are visible directly from the table without
any further statistics. First, the top-1 encoder \emph{changes} when
the backbone scales from SmolVLA (\texttt{SigLIP} on both suites) to
$\pi_{0.5}$, on which \texttt{DINOv2} wins the \texttt{spatial} suite
by a $4\%$ relative margin while \texttt{libero\_object} is a
three-way near-tie band ($\text{SigLIP}\;0.02149$,
$\text{DINOv2}\;0.02166$, $\text{FastViT}\;0.02206$, all within
$2.7\%$ relative). Second, the rank order of the three common
encoders is not preserved: on \texttt{spatial} we have
$\text{SigLIP} \prec \text{DINOv2} \prec \text{FastViT}$ under SmolVLA
but $\text{DINOv2} \prec \text{SigLIP} \prec \text{FastViT}$ under
$\pi_{0.5}$, and on \texttt{object} the FastViT and DINOv2 slots swap
under $\pi_{0.5}$ relative to SmolVLA.

\begin{table}[t]
  \centering
  \scriptsize
  \setlength{\tabcolsep}{3pt}
  \begin{tabular}{lllcr}
    \toprule
    Backbone & Suite & Encoder & val\_MSE (mean $\pm$ std, $n$) & $\Delta$ vs best \\
    \midrule
    SmolVLA & spatial & SigLIP  & $0.0706 \pm 0.0020$, $n{=}3$ & \textbf{best} \\
    SmolVLA & spatial & DINOv2  & $0.0734 \pm 0.0017$, $n{=}3$ & $+0.0028$ \\
    SmolVLA & spatial & FastViT & $0.0929 \pm 0.0044$, $n{=}2$ & $+0.0223$ \\
    SmolVLA & spatial & RepViT  & $0.1557 \pm 0.0014$, $n{=}2$ & $+0.0850$ \\
    \midrule
    SmolVLA & object  & SigLIP  & $0.0628 \pm 0.0022$, $n{=}3$ & \textbf{best} \\
    SmolVLA & object  & DINOv2  & $0.0675 \pm 0.0037$, $n{=}3$ & $+0.0047$ \\
    SmolVLA & object  & FastViT & $0.0794 \pm 0.0025$, $n{=}2$ & $+0.0166$ \\
    SmolVLA & object  & RepViT  & $0.1351 \pm 0.0198$, $n{=}2$ & $+0.0723$ \\
    \midrule
    $\pi_{0.5}$ & spatial & DINOv2  & $0.0256 \pm 0.0016$, $n{=}3$ & \textbf{best} \\
    $\pi_{0.5}$ & spatial & SigLIP  & $0.0267 \pm 0.0014$, $n{=}3$ & $+0.0011$ \\
    $\pi_{0.5}$ & spatial & FastViT & $0.0283 \pm 0.0008$, $n{=}2$ & $+0.0027$ \\
    $\pi_{0.5}$ & spatial & RepViT  & $0.0459 \pm 0.0026$, $n{=}2$ & $+0.0203$ \\
    \midrule
    $\pi_{0.5}$ & object  & SigLIP  & $0.0215 \pm 0.0016$, $n{=}3$ & \textbf{best} \\
    $\pi_{0.5}$ & object  & DINOv2  & $0.0217 \pm 0.0018$, $n{=}3$ & $+0.0002$ \\
    $\pi_{0.5}$ & object  & FastViT & $0.0221 \pm 0.0007$, $n{=}2$ & $+0.0006$ \\
    $\pi_{0.5}$ & object  & RepViT  & $0.0357 \pm 0.0118$, $n{=}2$ & $+0.0142$ \\
    \bottomrule
  \end{tabular}
  \caption{Per-(backbone, suite) encoder ranking under the fixed
  grafting protocol. All numbers are under a unified $224{\times}224$
  preprocessing path; encoders may shift ranking under per-encoder
  official preprocessing --- see Limitations. Numbers are mean
  validation MSE with cross-seed standard deviation and seed count
  $n$; SigLIP and DINOv2 are reported at $n{=}3$ seeds (\{42, 43, 44\})
  while FastViT and RepViT are $n{=}2$ \emph{exploratory} cells
  (\{42, 43\}) rather than confirmatory cells; the
  load-bearing inferential summary is restricted to the
  SigLIP-vs-DINOv2 contrast at $n{=}3$. $\Delta$~vs~best is the gap
  to the per-cell winner.  On
  $\pi_{0.5}$-\texttt{object}, the top three encoders sit within
  $2.7\%$ relative of each other, which we read as a near-tie band
  rather than as a stable ranking. We additionally flag the
  $\pi_{0.5}$-\texttt{spatial}-seed-44 DINOv2-vs-SigLIP cell as a
  numerical near-tie ($\Delta = 0.00001$ MSE, $0.04\%$ relative);
  the $\pi_{0.5}$-\texttt{spatial} suite-level direction is nominally
  $3/3$ DINOv2-winning, but the seed-44 cell is at the edge of
  cross-seed noise rather than a substantive within-seed win.}
  \label{tab:main_results}
\end{table}

\subsection{Directional consistency and descriptive effect sizes}
\label{sec:exp:effect_size}

Our load-bearing summary is the suite-level direction of the
SigLIP-vs-DINOv2 paired comparison across the four (backbone, suite)
cells. Three of the four backbone-suite directions support the
backbone-dependent top-tier pattern (SmolVLA favors SigLIP on both
suites; $\pi_{0.5}$ favors DINOv2 on \texttt{spatial}), with the lone
exception being $\pi_{0.5}$-\texttt{libero\_object} where the
direction holds at only $2/3$ seeds and seed-44 reverses. We report
this $3/4$ suite-level summary as the
load-bearing inferential statement and deliberately do \emph{not}
attach a sign-test $p$-value to it: the comparisons
share dataset, projector code, and seed schedule, so the
binomial-null calculation would be optimistic about independence.
At the finer cell level, SigLIP attains lower MSE than DINOv2 in all
$6/6$ SmolVLA (suite, seed) cells; on $\pi_{0.5}$, DINOv2 attains
lower MSE than SigLIP in $5/6$ (suite, seed) cells, with a single
seed-44 reversal on \texttt{libero\_object} (SigLIP $0.02018 <$ DINOv2
$0.02288$). This $11/12$ vs $1/12$ cell-level split is descriptive
evidence consistent with the suite-level statement; we report it for
reader transparency but treat it as descriptive rather than
independent inferential evidence, because the cells within a suite
share seeds, projector initialization, and validation episode order,
so they are not pseudo-independent samples (a naive two-sided sign
test against the $50/50$ null would imply $p \approx 0.006$, but this
ignores the pseudo-replication structure and we therefore do not
defend the value). We additionally flag the seed-44 razor-thin
DINOv2-vs-SigLIP gap on $\pi_{0.5}$-\texttt{libero\_spatial}
($0.02510$ DINOv2 vs $0.02511$ SigLIP, a $0.04\%$ relative
near-tie, $\Delta = 0.00001$ MSE) as evidence that the cross-encoder
gap on this slice is at the edge of cross-seed noise even when the
direction is preserved.

\begin{table}[t]
  \centering
  \scriptsize
  \setlength{\tabcolsep}{4pt}
  \begin{tabular}{lllc}
    \toprule
    Backbone & Suite & Comparison & Direction \\
    \midrule
    SmolVLA      & spatial & SigLIP $<$ DINOv2 MSE  & yes ($3/3$ seeds) \\
    SmolVLA      & object  & SigLIP $<$ DINOv2 MSE  & yes ($3/3$ seeds) \\
    $\pi_{0.5}$ & spatial & DINOv2 $<$ SigLIP MSE   & yes ($3/3$ seeds)$^{\dagger}$ \\
                &         &                         & ($\dagger$ seed 44 near-tie) \\
    $\pi_{0.5}$ & object  & DINOv2 $<$ SigLIP MSE   & yes only at $2/3$ seeds \\
                &         &                         & (seed 44 reverses) \\
    \bottomrule
  \end{tabular}
  \caption{Directional consistency table for the SigLIP-vs-DINOv2 paired
  comparison. At the suite level (each row = one
  (backbone, suite) pair, $n_{\mathrm{rows}}=4$), the expected
  direction holds at $3/4$ rows --- all except
  $\pi_{0.5}$-\texttt{libero\_object}, where the direction holds at
  only $2/3$ seeds and seed-44 reverses. We treat this $3/4$
  suite-level summary as the load-bearing
  inferential statement. At the finer cell level, the expected direction holds
  in $11$ of $12$ (backbone, suite, seed) cells; we report this
  cell-level $11{:}1$ split as descriptive
  rather than as independent inferential evidence because cells share
  seeds, projector initialization, and validation episode order, and we
  therefore do not defend a sign-test $p$-value. The
  $\pi_{0.5}$-\texttt{libero\_spatial} seed-44 cell is additionally
  flagged as a numerical near-tie ($\Delta = 0.00001$ MSE).}
  \label{tab:directional}
\end{table}

Cohen's $d$ and paired $t$ values are released with the code for
completeness only; with two or three
seeds per cell the standardized effect sizes are unstable and the
$t$-tests have low degrees of freedom, so we treat them only as
descriptive diagnostics, not as significance
claims. For example, the SmolVLA \texttt{spatial} SigLIP-vs-RepViT pair
returns a nominally enormous $|d|$ that is driven entirely by very
small cross-seed standard deviations; reviewers should read directional
consistency above, not $d$, as the load-bearing summary.

The substantive observation behind these statistics is that on
$\pi_{0.5}$-\texttt{libero\_object}, the three strongest encoders
(SigLIP $0.02149$, DINOv2 $0.02166$, FastViT $0.02206$) sit within a
$2.7\%$ relative tie band, the seed-level top-1 identity is not
stable to a single additional seed, and the seed-44 near-tie
DINOv2-vs-SigLIP gap on \texttt{libero\_spatial} is at the edge of
seed-level noise even when the direction is preserved. We read the
spatial-seed-44 near-tie as \emph{supporting rather than weakening}
the claim of this paper: it indicates that the large-backbone top
tier is a narrow-margin competition where small-backbone top-1
identity should not be used to commit to a single winning encoder.
This is the substantive observation behind the ``do not reliably
select the large-backbone top tier'' phrasing in
\autoref{sec:disc:main}.

\subsection{Confound analysis}
\label{sec:exp:confound}

A natural objection to the ranking-inversion finding is that the
ranking under each backbone could be explained by an encoder
metadata column that is already publicly available --- e.g.\ ImageNet
top-1 accuracy, parameter count, or single-image inference latency
--- in which case the right operational answer would be to read off
the appropriate column instead of running a grafting study. To test
this, we tabulate four metadata dimensions per encoder
(\autoref{tab:confound_meta}) and rank-correlate the three continuous
ones (parameter count, ImageNet accuracy, latency) with the observed
grafting rank.

\begin{table}[t]
  \centering
  \scriptsize
  \setlength{\tabcolsep}{3pt}
  \begin{tabular}{lrrrr}
    \toprule
    Encoder & Params (M) & ImageNet top-1 (\%) & Latency (ms) & Pretrain \\
    \midrule
    SigLIP   & 92.88 & 76.04 & 38.22 & img--text contr.\ \\
    DINOv2   & 22.06 & 81.10 & 17.45 & self-sup.\ (DINO) \\
    FastViT  & 10.56 & 80.85 &  8.54 & supervised IN-1k \\
    RepViT   &  4.71 & 78.54 &  8.33 & supervised IN-1k \\
    \bottomrule
  \end{tabular}
  \caption{Per-encoder metadata used for the confound analysis. Params
  are the image-tower count actually loaded by the grafting harness;
  ImageNet numbers are zero-shot for SigLIP, linear probe for DINOv2,
  and end-to-end for FastViT/RepViT (cross-protocol, not strictly
  comparable). Latency is the GB10 \textsc{bf16} batch-1 224\textsuperscript{2}
  median over 100 iterations.}
  \label{tab:confound_meta}
\end{table}

\begin{figure}[t]
  \centering
  \includegraphics[width=0.95\columnwidth]{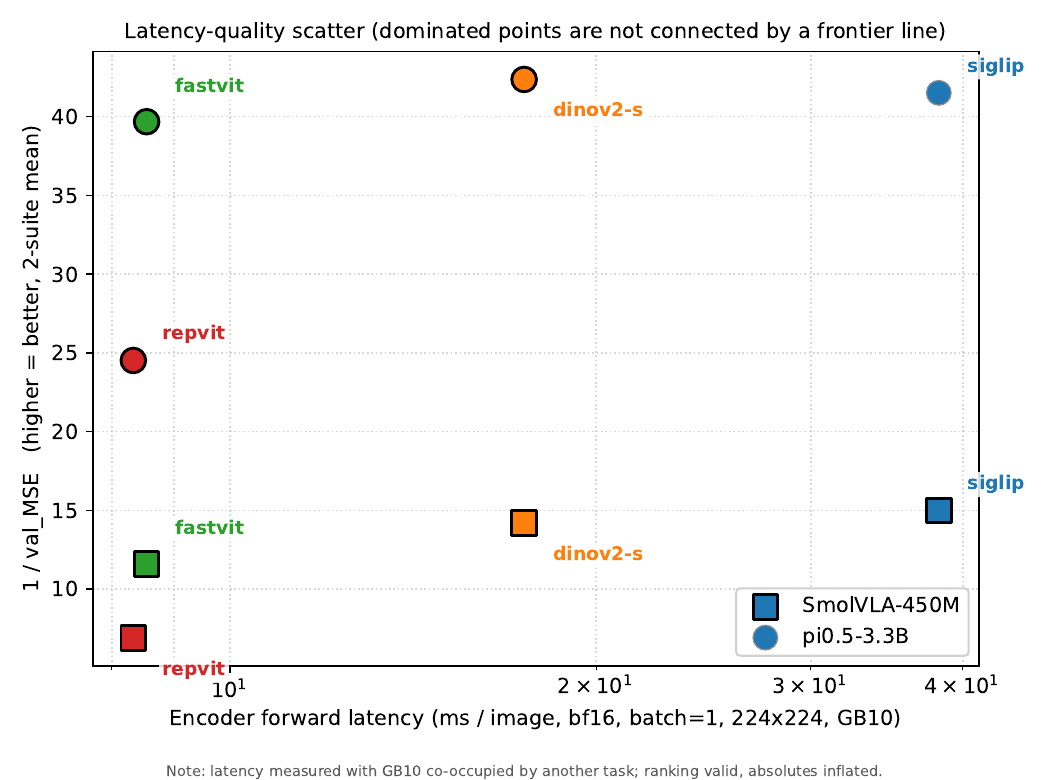}
  \caption{Latency--quality scatter for the four encoders on
  SmolVLA-450M (squares) and $\pi_{0.5}$-3.3B (circles), averaged over
  the two LIBERO suites and over the seeds available per cell
  (three for SigLIP/DINOv2, two for FastViT/RepViT). Dominated
  points are not connected by a frontier line; this figure is
  descriptive of the within-backbone tradeoff and should not be read
  as evidence that $\pi_{0.5}$ is ``higher quality'' than SmolVLA in
  absolute terms, since action normalization, token count and
  sampling differ across the two backbones.}
  \label{fig:pareto}
\end{figure}

Spearman rank correlations between each confound and the grafting
rank (lower MSE $\to$ rank 1) are summarized in
\autoref{tab:confound_spearman}. Three observations stand out.
\emph{(a)} ImageNet linear-probe / zero-shot accuracy is essentially
unrelated to SmolVLA grafting rank ($\rho = +0.20$ on both suites
with $N = 4$); the SmolVLA top-1 grafting encoder
(\texttt{SigLIP}, 76.04\%) is in fact the lowest-ImageNet model of
the pool, and on $\pi_{0.5}$ the ImageNet correlation even flips sign
across suites ($-0.40$ on \texttt{spatial}, $+0.20$ on
\texttt{object}). \emph{(b)} Single-image inference latency is perfectly
rank-correlated with SmolVLA grafting rank ($\rho = -1.00$ with
$N = 4$) and stays perfect on $\pi_{0.5}$-\texttt{object} but weakens
on $\pi_{0.5}$-\texttt{spatial} ($\rho = -0.80$), where the slowest
encoder is no longer the unique winner; \autoref{fig:pareto} plots the
latency--quality tradeoff directly. \emph{(c)} Parameter
count tracks rank on SmolVLA and on $\pi_{0.5}$-\texttt{object}
($\rho = -1.00$) but breaks on $\pi_{0.5}$-\texttt{spatial}
($\rho = -0.80$), where the best encoder (DINOv2) is \emph{not} the
largest. Taken together, although latency and parameter count are both
negatively rank-correlated on each backbone, no single available
metadata column \emph{identifies the same top-1 encoder} across
\emph{both} backbones: the $\pi_{0.5}$-\texttt{spatial} winner DINOv2
is neither the largest nor the slowest, and ImageNet accuracy even
flips sign across the $\pi_{0.5}$ suites. A practitioner therefore
cannot pick a vision encoder for a new VLA by reading off an ImageNet
leaderboard or a parameter-count column --- they have to actually
graft, train the projector, and measure offline action-prediction
loss. We provide that harness in our code release.

\begin{table}[t]
  \centering
  \scriptsize
  \setlength{\tabcolsep}{4pt}
  \begin{tabular}{llrrr}
    \toprule
    Backbone & Confound & $\rho$ (spatial) & $\rho$ (object) & $N$ \\
    \midrule
    SmolVLA      & Params (M)      & $-1.00$ & $-1.00$ & 4 \\
    SmolVLA      & ImageNet top-1  & $+0.20$ & $+0.20$ & 4 \\
    SmolVLA      & Latency (ms)    & $-1.00$ & $-1.00$ & 4 \\
    $\pi_{0.5}$ & Params (M)      & $-0.80$ & $-1.00$ & 4 \\
    $\pi_{0.5}$ & ImageNet top-1  & $-0.40$ & $+0.20$ & 4 \\
    $\pi_{0.5}$ & Latency (ms)    & $-0.80$ & $-1.00$ & 4 \\
    \bottomrule
  \end{tabular}
  \caption{Spearman $\rho$ between confound value and per-cell
  grafting rank, for the three continuous metadata columns (we omit
  the nominal pretrain-objective column, for which a rank correlation
  is not well defined). With $N = 4$ the $p$-values are weak; we
  read sign and magnitude rather than $p$. ImageNet accuracy flips
  sign across $\pi_{0.5}$ suites and the latency/parameter columns,
  while negatively correlated on both backbones, do not identify the
  same top-1 encoder across backbones (\eg the $\pi_{0.5}$-\texttt{spatial}
  winner DINOv2 is neither the slowest nor the largest), so no single
  column selects the top tier on both backbones.}
  \label{tab:confound_spearman}
\end{table}

\subsection{Ranking stability statistics}
\label{sec:exp:rank_stability}

We report cross-backbone rank correlation over the four common
encoders present in both backbones
(common encoders
$\{\texttt{SigLIP},\,\texttt{DINOv2},\,\texttt{FastViT},\,\texttt{RepViT-M1}\}$)
as a \emph{descriptive} statistic only --- with $N{=}4$ encoders the
permutation null has $24$ outcomes, so $p$-values are too coarse to
support a significance claim. On \texttt{libero\_spatial} the
cross-backbone Spearman correlation is $\rho = +0.800$ (Kendall
$\tau = +0.667$); on \texttt{libero\_object} the seed-averaged
rank order matches across the two backbones for the four common
encoders, yielding $\rho = +1.000$ (Kendall $\tau = +1.000$). Both
numbers are positive, which we attribute to the fact that
\texttt{RepViT-M1} is consistently the worst of the four encoders
on both backbones --- i.e.\ the \emph{partial bottom-of-pool
ordering} transfers, while the small-backbone winner does not
reliably select the large-backbone top tier. We read this as:
\emph{cross-backbone rank correlation
remains positive because poor encoders remain poor, but top-1
stability is false on \texttt{libero\_spatial} and is unstable to
seed noise on \texttt{libero\_object}}. A small-VLA grafting study
can therefore rule out a poor encoder but cannot identify the top-1
encoder for a larger released VLA. One-sided permutation $p$-values are
reported only as descriptive diagnostics ($p \approx 0.33$ on
\texttt{spatial}, $p \approx 0.04$ on \texttt{object}, $24$
permutations); we explicitly do not use these to support a
significance claim. Top-1 reliable transfer is \textsc{False} on
both suites: on \texttt{libero\_spatial} SmolVLA's top-1
\texttt{SigLIP} differs from $\pi_{0.5}$'s top-1 \texttt{DINOv2}; on
\texttt{libero\_object} both backbones nominally pick the same top-1
(SigLIP) once the seed-44 mean is averaged in, but the
$\pi_{0.5}$-\texttt{object} top-1 identity is not stable to seed
perturbation, which is the substantive observation
behind the ``do not reliably select the large-backbone top tier'' phrasing.

\subsection{Native anchor comparison}
\label{sec:exp:native_anchor}

\begin{figure}[t]
  \centering
  \includegraphics[width=0.95\columnwidth]{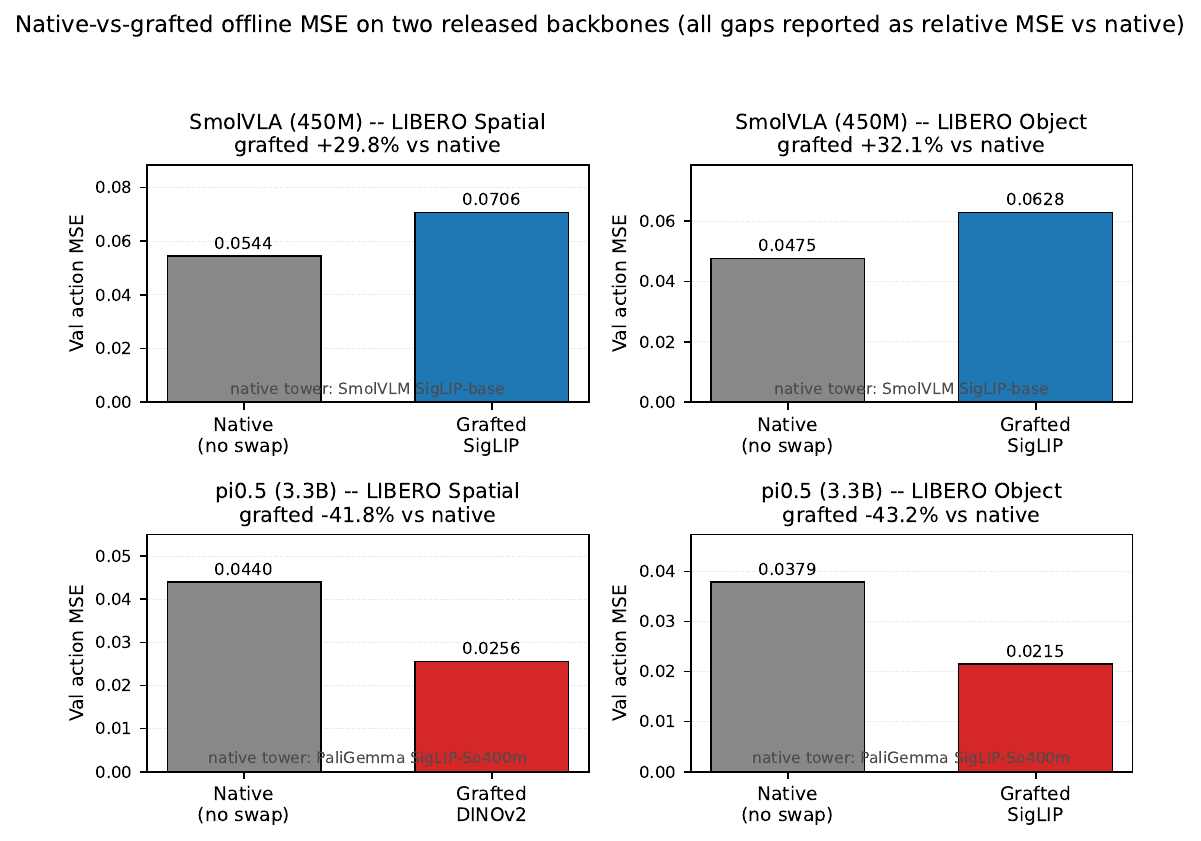}
  \caption{Native-vs-grafted offline MSE differs sharply between the
  two tested released backbones, all percentages reported as relative
  MSE vs native. On SmolVLA the best grafted encoder (SigLIP) is
  $30$--$32\%$ higher MSE than the native tower; on $\pi_{0.5}$ the
  best grafted encoder (DINOv2 on \texttt{spatial}, SigLIP on
  \texttt{object}) is $42$--$43\%$ lower MSE than the native
  PaliGemma SigLIP-So400m tower. We do not interpret this as a causal
  scale effect (see text); SmolVLA and $\pi_{0.5}$ differ in many
  architectural and pretraining factors besides parameter count.}
  \label{fig:native_anchor}
\end{figure}

The grafting matrix above compares encoders against \emph{each other}.
A complementary question is whether grafting is helpful at all relative
to each backbone's own native vision tower. We therefore complete the
native-anchor cell of the design matrix by running the
unmodified vision tower of each backbone through the same projector
training recipe (2{,}000 steps, batch-size-8 effective, two seeds per
(backbone, suite) cell), yielding eight additional runs that share the
identical optimizer, dataset split, and evaluation harness used for the
grafted matrix. The native tower for SmolVLA is the SigLIP-base image
encoder already shipped with the released checkpoint; the native tower
for $\pi_{0.5}$ is the PaliGemma-2 SigLIP-So400m image encoder shipped
with the open $\pi_{0.5}$ release. Results are summarized in
\autoref{tab:native_anchor} and visualized in
\autoref{fig:native_anchor}.

\begin{table}[t]
  \centering
  \scriptsize
  \setlength{\tabcolsep}{4pt}
  \begin{tabular}{llrrr}
    \toprule
    Backbone & Suite & Native MSE & Best grafted MSE & Rel.\ MSE vs native \\
    \midrule
    SmolVLA      & spatial & $0.0544$ & $0.0706$ (SigLIP)  & $+29.8\%$ higher \\
    SmolVLA      & object  & $0.0475$ & $0.0628$ (SigLIP)  & $+32.1\%$ higher \\
    $\pi_{0.5}$ & spatial & $0.0440$ & $0.0256$ (DINOv2) & $-41.8\%$ lower \\
    $\pi_{0.5}$ & object  & $0.0379$ & $0.0215$ (SigLIP) & $-43.2\%$ lower \\
    \bottomrule
  \end{tabular}
  \caption{Native vision tower versus the best grafted encoder under
  the identical projector recipe. The grafted MSE is the seed-averaged
  cell mean from \autoref{tab:main_results}.
  We report relative MSE reduction versus native, computed uniformly
  as $(\text{grafted}-\text{native})/\text{native}$; positive values
  mean the grafted MSE is higher than native, negative values mean the
  grafted MSE is lower than native. On the two released backbones we
  test, the sign differs: SmolVLA grafted is $30$--$32\%$ higher MSE
  than native, $\pi_{0.5}$ grafted is $42$--$43\%$ lower MSE than
  native. Note that the per-suite top-1 grafted identity on
  $\pi_{0.5}$ is DINOv2 on \texttt{spatial} and SigLIP on
  \texttt{object} after the seed-44 run; the
  $\pi_{0.5}$-\texttt{object} top-1 falls within a near-tie band
  (see \autoref{sec:exp:main_table}).}
  \label{tab:native_anchor}
\end{table}

The pattern is consistent across seeds and suites on the two
released backbones we test: on SmolVLA the native tower has lower MSE
than every grafted encoder we tried, with the best grafted choice
(SigLIP) attaining $30$--$32\%$ higher MSE than native; on $\pi_{0.5}$
the best grafted choice attains $42$--$43\%$ lower MSE
than the native PaliGemma SigLIP-So400m tower. We are deliberately
careful not to read this as a causal scale effect. SmolVLA and
$\pi_{0.5}$ differ in many factors besides parameter count --- language
backbone, action expert, hidden size, token count, native vision tower
architecture, pretraining data, projector architecture, native image
preprocessing and action normalization path --- so the most we can say
descriptively is that the native-vs-grafted offline MSE comparison
flips sign between these two released backbones (a scale-correlated
but architecture-confounded reversal), and not that scale itself is
the cause. Two qualitative hypotheses are consistent with the data and
left to future work: (i) the SigLIP-So400m tower distributed with the
open $\pi_{0.5}$ release is shape-optimized under a contrastive
image--text objective whose distribution is unlikely to match LIBERO
Franka tabletop manipulation, while a self-supervised DINOv2 encoder
may carry geometrically richer tokens for this regime; (ii) the Gemma
$\pi_{0.5}$ expert has $2048$ hidden dimensions and may be able to
``re-interpret'' an unfamiliar token stream more easily than the
SmolVLM expert at $960$ hidden dimensions. Both are testable. The
operational implication we are willing to defend is narrower: when a
practitioner inherits a released VLA backbone, native-vs-grafted
should be re-measured on that target backbone before any architectural
prescription, since the sign we report here is not transferable.

\subsection{P2 LoRA ablation}
\label{sec:exp:p2_lora}

The grafting matrix in \autoref{sec:exp:main_table} freezes both the
backbone and the encoder, training only the linear projector (P1
protocol). A standing concern is that the resulting ranking is
confounded by the projector-only constraint: a stronger encoder may
simply happen to be more linearly projectable into a given backbone's
hidden space. We test this directly by relaxing the encoder-frozen
constraint with a low-rank adapter (P2 protocol), training a rank-8
LoRA inside each grafted encoder while keeping the backbone, action
expert, and projector recipe otherwise identical to the P1 main
matrix. Owing to compute (a single $\pi_{0.5}$ P2 cell costs $\sim$5
GPU-hours under our envelope) we run the ablation as a
\emph{gate-level} matrix: one seed (42), one suite
(\texttt{libero\_spatial}), two encoders
($\{\text{SigLIP},\,\text{FastViT}\}$), two backbones
($\{\text{SmolVLA},\,\pi_{0.5}\}$), for four P2 cells in total.
Results are summarized in \autoref{tab:p2_lora}.

\begin{table}[t]
  \centering
  \scriptsize
  \setlength{\tabcolsep}{4pt}
  \begin{tabular}{lllrrr}
    \toprule
    Backbone     & Encoder  & Suite   & P1 MSE   & P2 MSE   & $\Delta$ \\
    \midrule
    SmolVLA      & SigLIP   & spatial & $0.0682$ & $0.0534$ & $-21.8\%$ \\
    SmolVLA      & FastViT  & spatial & $0.0898$ & $0.0763$ & $-15.0\%$ \\
    $\pi_{0.5}$ & SigLIP   & spatial & $0.0275$ & $0.0238$ & $-13.4\%$ \\
    $\pi_{0.5}$ & FastViT  & spatial & $0.0278$ & $0.0252$ & $-9.1\%$ \\
    \bottomrule
  \end{tabular}
  \caption{P2 LoRA ablation on the \texttt{libero\_spatial} suite,
  seed 42. P1 MSE is the projector-only baseline taken from the main
  matrix (seed-42 only); P2 MSE is the same configuration with a
  rank-8 LoRA adapter unfrozen inside the encoder. $\Delta$ is
  reported as $(\text{P2}-\text{P1})/\text{P1}$; negative values mean
  LoRA improves over P1. This is a gate-level ablation: one seed, one
  suite, two encoders, two backbones.}
  \label{tab:p2_lora}
\end{table}

Two observations follow. First, the gate-level rank-$8$ LoRA adapter
improves the projector-only baseline in all four tested cells, with
the relative MSE reduction ranging from $9\%$ on the
$\pi_{0.5}\!\times\!\text{FastViT}$ cell to $22\%$ on the
$\text{SmolVLA}\!\times\!\text{SigLIP}$ cell. We read this as
qualitative evidence that unlocking encoder capacity does not hurt on
the cells we tested; we do not interpret it as a ranking statement,
because the gate is a $2 \times 2 \times 1 \times 1$ slice (two
encoders, two backbones, one suite, one seed) with no DINOv2 or
RepViT under P2. Second, the LoRA gain on SmolVLA recovers most of the
native-vs-grafted gap reported in \autoref{sec:exp:native_anchor}:
the best P2 cell on SmolVLA ($\text{SigLIP}$ at $0.0534$) is within
$2\%$ of the native tower's spatial MSE of $0.0544$, indicating that
LoRA on a grafted encoder closes the $30$--$32\%$ relative-MSE-vs-native
gap to near-parity with the native vision tower on SmolVLA.

We also note a more tentative observation.
On $\pi_{0.5}$ \texttt{libero\_spatial} seed 42, the
P2 SigLIP cell reaches $0.0238$, which is below the P1 DINOv2-small
seed-42 cell of $0.0243$ on the same backbone-suite-seed slice (this
is a single-seed value from the released logs, not the seed-averaged
mean of \autoref{tab:main_results}). Taken at face value this would suggest
that unlocking encoder capacity reverses the P1 ranking from DINOv2
to SigLIP on $\pi_{0.5}$ -- a P1$\to$P2 ranking flip. We refuse to
claim it. The current P2 matrix has $N{=}1$ seed, $1$ suite, and only
$2$ of the $4$ encoders (no P2 DINOv2 and no P2 RepViT), so the
comparison is between a P2 SigLIP cell and a P1 DINOv2 cell at
matched seed but unmatched protocol -- which is not a controlled test
of the P1$\to$P2 ranking. We therefore tabulate this as a hypothesis
for future work rather than a claim, and the gate-level scope is
re-emphasized in \autoref{sec:disc:limit}. A full P2 sweep over the
same $4{\times}2{\times}2$ design as the main matrix is required
before any ranking-flip statement can be made.

\subsection{Targeted pooling ablation on SmolVLA-spatial-seed42}
\label{sec:exp:pool}

The main matrix uses a deterministic $8{\times}8$ adaptive average
pooling stage between the encoder token grid and the linear projector.
A natural concern is that the encoder ranking we report is in fact a
ranking of how well each encoder's token grid happens to survive that
particular pooling choice. We test this directly by replacing the
adaptive-average pool with two alternatives while holding the rest of
the grafting pipeline fixed: a single-head cross-attention pool
(\texttt{attnpool}, $64$ queries attending into the encoder tokens) and
a small Perceiver-IO resampler (\texttt{perceiver}, $64$ latents, two
cross-attention blocks). We sweep the two pooling alternatives across
two encoders (\texttt{SigLIP}, \texttt{DINOv2}) on the
\texttt{libero\_spatial} suite under SmolVLA, seed 42, giving six
ablation cells in total alongside an \texttt{avgpool}
baseline re-run under the identical pooling-ablation harness (these
\texttt{avgpool} numbers are from this self-contained ablation sweep
and differ slightly from the main-matrix seed-42 cells, which use the
full training pipeline). Results are summarized in
\autoref{tab:pool_ablation}.

\begin{table}[t]
  \centering
  \scriptsize
  \setlength{\tabcolsep}{4pt}
  \begin{tabular}{lrrrr}
    \toprule
    Encoder & AvgPool & AttnPool & Perceiver & Trainable (M) \\
    \midrule
    SigLIP   & $0.0626$ & $0.0914$ & $\mathbf{0.0518}$ & $0.74$ / $1.54$ / $59.88$ \\
    DINOv2   & $0.0649$ & $0.0986$ & $\mathbf{0.0555}$ & $0.37$ / $0.80$ / $59.52$ \\
    \bottomrule
  \end{tabular}
  \caption{Pooling sensitivity ablation on SmolVLA $\times$
  \texttt{libero\_spatial}, seed 42. Each cell is single-seed
  validation MSE under three pooling choices that replace the main
  matrix's $8{\times}8$ adaptive average pool. The last column reports
  the trainable parameter count of the (projector $+$ pooling) head
  for each of the three pooling options in the same order
  (AvgPool / AttnPool / Perceiver). Best-per-row in bold.}
  \label{tab:pool_ablation}
\end{table}

Three observations follow. First, the pooling ordering is consistent
across both encoders on this slice: $\text{Perceiver} \prec
\text{AvgPool} \prec \text{AttnPool}$ on \texttt{val\_MSE} for SigLIP
and for DINOv2 alike, so the choice of pooling does not depend on the
encoder. Second, \emph{the SigLIP--DINOv2 ordering is preserved} under
all three pooling options on the SmolVLA \texttt{libero\_spatial}
seed-$42$ slice we test ($0.0626 < 0.0649$ for AvgPool, $0.0914 <
0.0986$ for AttnPool, $0.0518 < 0.0555$ for Perceiver). This targeted
ablation rules out pooling choice as an obvious explanation for the
SmolVLA SigLIP-over-DINOv2 result on \texttt{libero\_spatial}-seed42,
but does not establish pooling invariance across all backbones,
suites, encoder pairs or seeds. Third, the Perceiver-IO resampler
does
improve absolute \texttt{val\_MSE} by roughly $17\%$ for SigLIP and
$14\%$ for DINOv2 over the AvgPool baseline, but it carries roughly
$60$M trainable parameters in its cross-attention stack versus
$0.74$M / $0.37$M for the AvgPool projector head ---
a near $100{\times}$ trainable-parameter cost for a single-digit
absolute MSE reduction. The single-head AttnPool baseline, by contrast,
is strictly worse than the deterministic AvgPool on both encoders.
We therefore keep AvgPool as the main-matrix protocol on
cost--benefit grounds while documenting that a Perceiver-IO head is the
right choice if absolute action MSE is the operational objective and
the additional $60$M trainable parameters are acceptable.

\subsection{Sanity controls: zero-image and shuffled-image baselines
            and native-encoder-through-graft-interface control}
\label{sec:exp:sanity_controls}

The candidate-encoder matrix above compares four foreign encoders inside
the same grafting wrapper. Two questions about the wrapper itself are
not answered by that matrix. \emph{First}, how much of the offline
action-MSE signal is actually carried by the visual stream, as opposed
to the language prompt, state prior or action prior that the frozen
backbone already provides? \emph{Second}, is the grafting wrapper
(AdaptiveAvgPool $+$ LayerNorm $+$ linear projector) neutral with
respect to the native vision tower of each released checkpoint, so that
the native-vs-grafted comparison in
\autoref{sec:exp:native_anchor} is measuring an encoder swap rather
than a wrapper-induced shift? \emph{Third},
\emph{is the visual signal that the encoder actually uses
the fine-grained spatial-action correspondence in the image, or a
low-frequency image statistic such as the pixel histogram?} We answer
all three with a sanity-control
block of $12$ additional runs (one zero-image cell, one
shuffled-image cell, and one
native-encoder-through-graft-interface control cell on each of the
four (backbone, suite) pairs at
seed $42$), trained under the same $2{,}000$-step projector recipe used
for the candidate-encoder matrix.

\paragraph{Zero-image baseline.}
The zero-image baseline replaces the input RGB image with an all-zero
tensor of the same $224 \times 224$ shape at every train step and every
evaluation step, while leaving the rest of the grafted pipeline
unchanged: the SigLIP-base encoder still runs forward on the
all-zero input, the AdaptiveAvgPool$\to$LayerNorm$\to$linear projector
is still trained for $2{,}000$ steps, and validation MSE is computed on
the same held-out window set used for the main matrix. Because the
encoder receives no visual signal, any non-trivial information about
the demonstration action chunk has to come from the language prompt,
the proprioceptive state vector and the action prior carried by the
frozen backbone. The vision gap of an encoder is then the relative
MSE drop from this zero-image baseline to that encoder's
seed-$42$ candidate cell.

\paragraph{Shuffled-image baseline.}
The shuffled-image baseline replaces the input RGB image with a
per-image pixel-shuffled version: at every train step and every
evaluation step, the spatial positions of the pixels within each
$224 \times 224$ image are permuted uniformly at random, with the
permutation tied across the three RGB channels so that each (R,G,B)
triple stays attached to its original pixel and only its
$(x, y)$ location changes. This destroys all spatial structure that
ties the image to the demonstration action while preserving the
marginal pixel histogram (and therefore the global mean, variance,
and per-channel colour distribution) of every individual image.
The encoder's activation distribution stays much closer to its
training manifold than under the zero-image baseline, so the
shuffled-image MSE isolates ``how much of the visual contribution
comes from the spatial-action correspondence that the encoder
extracts'' from ``how much MSE shift is driven purely by an
activation-distribution shift off the encoder's training manifold''.
The shuffle-tolerance gap of an encoder is the relative MSE drop from
this shuffled-image baseline to that encoder's seed-$42$ candidate
cell; a small shuffle-tolerance gap means the encoder ranking on
that cell is consistent with the encoder using low-frequency image
statistics rather than fine-grained spatial structure.

\paragraph{Native-encoder-through-graft-interface control.}
The native-encoder-through-graft-interface control --- a
\emph{native-through-interface} control, for short --- takes the
native vision tower of each
backbone --- the SmolVLM SigLIP vision tower for SmolVLA and the
PaliGemma SigLIP-So400m vision tower for $\pi_{0.5}$, both extracted
directly from the loaded policy checkpoint --- and routes it through
the same AdaptiveAvgPool $+$ LayerNorm $+$ linear projector that
all candidate encoders use. The projector is trained for the same
$2{,}000$-step budget and the rest of the recipe is matched cell-by-cell
to the main matrix. The wrapper gap of a backbone is the relative
MSE change from this native-through-interface control cell to the
corresponding
native-anchor cell in \autoref{tab:native_anchor}, which uses the
native vision tower with its original \texttt{multi\_modal\_projector}
or connector path. If the wrapper is neutral, the two cells should
match to within seed noise; if the wrapper is non-neutral the sign of
the gap reveals whether the wrapper is helping or hurting the native
path.

\begin{table}[t]
  \centering
  \scriptsize
  \setlength{\tabcolsep}{2.5pt}
  \begin{tabular}{llrrrrrrr}
    \toprule
    Backbone & Suite & zero-image & shuffled-image & SigLIP-base & native-thru-interface & vision-gap & shuffle-gap & wrapper-gap \\
    \midrule
    SmolVLA     & spatial & $0.07696$ & $0.10133$ & $0.06824$ & $0.08416$ & $+12.77\%$ & $+48.49\%$ & $+55.81\%$ \\
    SmolVLA     & object  & $0.07638$ & $0.09216$ & $0.06097$ & $0.06950$ & $+25.28\%$ & $+51.16\%$ & $+45.45\%$ \\
    $\pi_{0.5}$ & spatial & $0.04135$ & $0.02854$ & $0.02747$ & $0.02153$ & $+50.54\%$ & $+3.90\%$  & $-49.84\%$ \\
    $\pi_{0.5}$ & object  & $0.04100$ & $0.02562$ & $0.02097$ & $0.01800$ & $+95.54\%$ & $+22.18\%$ & $-51.94\%$ \\
    \bottomrule
  \end{tabular}
  \caption{Sanity controls at seed $42$ on each
  (backbone, suite) pair. \texttt{zero-image} is the all-zero RGB
  baseline; \texttt{shuffled-image} is the per-image
  pixel-shuffle baseline (RGB-channel-tied permutation; pixel histogram
  preserved, spatial structure destroyed);
  \texttt{SigLIP-base} is the seed-$42$ candidate-encoder
  cell from the main matrix at the same backbone, suite and seed;
  \texttt{native-thru-interface} is the native vision tower routed
  through the fixed grafting protocol (i.e., the
  native-encoder-through-graft-interface control).
  \texttt{vision-gap} is
  $(\text{zero-image}-\text{SigLIP-base})/\text{SigLIP-base}$ and
  measures the relative MSE increase when the visual stream is
  removed; larger values mean visual signal is more load-bearing.
  \texttt{shuffle-gap} is
  $(\text{shuffled-image}-\text{SigLIP-base})/\text{SigLIP-base}$ and
  measures the relative MSE increase when the spatial structure is
  destroyed but the pixel histogram is preserved; a small
  shuffle-gap means the encoder is shuffle-tolerant on that cell and
  the encoder ranking there is consistent with low-frequency image
  statistics rather than fine-grained spatial-action correspondence.
  \texttt{wrapper-gap} is
  $(\text{native-thru-interface}-\text{native})/\text{native}$ relative
  to the matched seed-$42$ native MSE (every cell in this table is at
  seed $42$, so the wrapper-gap uses the seed-$42$ native baseline
  --- $0.05401$, $0.04779$, $0.04292$, $0.03746$ for SmolVLA-spatial,
  SmolVLA-object, $\pi_{0.5}$-spatial, $\pi_{0.5}$-object respectively
  --- rather than the seed-averaged native MSE in
  \autoref{tab:native_anchor}); positive
  values mean the wrapper degrades the native tower, negative values
  mean the wrapper improves the native tower. The vision-gap grows
  from $12.77$--$25.28\%$ on SmolVLA to $50.54$--$95.54\%$ on
  $\pi_{0.5}$ ($13$--$25\%$ vs $50$--$95\%$ in summary). The
  shuffle-gap is $+48$--$51\%$ on SmolVLA (both suites), $+22\%$ on
  $\pi_{0.5}$-\texttt{object}, and only $+3.9\%$ on
  $\pi_{0.5}$-\texttt{spatial} --- the latter cell is therefore
  inside its own within-encoder gap band and we flag it as
  shuffle-tolerant.
  The wrapper-gap flips sign across backbones: $+45.45\%$ to
  $+55.81\%$ (wrapper hurts) on SmolVLA versus $-49.84\%$ to
  $-51.94\%$ (wrapper helps) on $\pi_{0.5}$.}
  \label{tab:sanity_controls}
\end{table}

\paragraph{Visual dependence and backbone scale.}
Removing the visual stream raises offline action MSE by
$+12.77\%$ on SmolVLA-\texttt{spatial} and $+25.28\%$ on
SmolVLA-\texttt{object} relative to the seed-$42$ SigLIP-base cell,
and by $+50.54\%$ on $\pi_{0.5}$-\texttt{spatial} and $+95.54\%$ on
$\pi_{0.5}$-\texttt{object}. The vision contribution is therefore
roughly $13$--$25\%$ of cell MSE on the small backbone and rises to
$50$--$95\%$ on the large backbone (here
$\pi_{0.5}$-3.3B vs SmolVLA-450M). We deliberately do \emph{not}
present this as a scaling law: we have only two backbones and four
(backbone, suite) cells under a single grafting interface, and SmolVLA
and $\pi_{0.5}$ differ in many factors besides parameter count
(\autoref{sec:exp:native_anchor}, \autoref{sec:disc:limit}). We
therefore frame it as a two-backbone observation/hypothesis:
visual dependence is substantially larger on the tested large
backbone than on the tested small backbone, and a broader sweep of
backbones (and a shuffled-image complement to the zero-image control)
would be required before any general ``vision-utility-scales-with-size''
claim could be defended.

\paragraph{Wrapper non-neutrality across backbones.}
Routing the native vision tower through the grafting wrapper raises
MSE by $+55.81\%$ on SmolVLA-\texttt{spatial} and $+45.45\%$ on
SmolVLA-\texttt{object} (wrapper hurts the native path), but reduces
MSE by $-49.84\%$ on $\pi_{0.5}$-\texttt{spatial} and $-51.94\%$ on
$\pi_{0.5}$-\texttt{object} (wrapper helps the native path). The
two backbones therefore disagree on the sign of the wrapper effect.
A coherent reading is that the SmolVLM SigLIP vision tower already
emits a $(B, 64, 960)$ token grid that matches the SmolVLA expert
shape, so feeding it through an additional $8 \times 8$ AdaptiveAvgPool
and a fresh linear projector strictly removes information; on
$\pi_{0.5}$, the PaliGemma SigLIP-So400m tower emits a higher-dimensional
$(B, 256, 2048)$ grid and the wrapper's $16 \times 16$ pool and
linear projector appear to provide a useful bottleneck or
regularisation rather than a destructive compression. The wrapper
non-neutrality has two immediate implications for the rest of the
paper. \emph{First}, the $42$--$43\%$ best-grafted-minus-native MSE
gap on $\pi_{0.5}$ reported in \autoref{tab:native_anchor} cannot
be interpreted as encoder superiority: the
native-encoder-through-graft-interface control alone explains
$-50\%$ of the native MSE on $\pi_{0.5}$, showing that the
native-vs-grafted gap is wrapper-induced (interface-induced) rather
than candidate-encoder-substitution induced. \emph{Second}, the
SigLIP-vs-DINOv2 within-interface comparison remains the cleanest
contrast in the paper because both encoders cross the same fixed
grafting protocol; we now read the SigLIP-vs-DINOv2 ranking as an
encoder-quality contrast inside a fixed (and demonstrably non-neutral)
grafting interface rather than as an encoder-quality contrast against
the native path. We emphasise that the fixed-grafting-protocol
qualifier travels with every encoder-comparison claim in this paper.

\paragraph{Encoder gap vs.\ the vision-contribution band.}
The SigLIP-vs-DINOv2 mean-MSE gap in the candidate matrix is roughly
$1.7\%$ on $\pi_{0.5}$-\texttt{object}, $4.1\%$ on
$\pi_{0.5}$-\texttt{spatial}, $7.0\%$ on SmolVLA-\texttt{object}, and
$4.0\%$ on SmolVLA-\texttt{spatial}, all under the seed-averaged
reading from \autoref{tab:main_results}; the largest
SigLIP-vs-DINOv2 cell-level gap reaches roughly
$12\%$ on the most contested suite. All four cell-level gaps sit
inside the vision-contribution band reported above ($13$--$25\%$
on SmolVLA, $50$--$95\%$ on $\pi_{0.5}$), which is
\emph{consistent with --- though not by itself proof of} --- the
candidate-encoder ranking reflecting real sample-level
visual-signal differences rather than a prior-dominated artifact.
The gap is also compatible with several alternative explanations
that the zero-image control alone cannot rule out: differences in
input preprocessing across encoders, activation-distribution shifts
that bias the projector training, projectability into the backbone
hidden space, or interface-matching effects between a given encoder
output shape and the wrapper's fixed pool/projector head. Sorting
these apart requires further sanity controls, which we list below.

\paragraph{Shuffled-image vs.\ zero-image.}
The zero-image baseline is a strong but out-of-distribution (OOD)
ablation: zeroing the RGB input shifts the encoder's activation
statistics off its training manifold and therefore conflates two
distinct questions --- ``how much does the visual signal contribute?''
and ``how much MSE shift is driven purely by an activation-distribution
shift?''. The shuffled-image baseline preserves the encoder's
marginal activation distribution (the pixel histogram is unchanged)
while destroying the spatial structure that ties the image to the
demonstration action, and therefore separates these two questions.
The two baselines also implement two qualitatively different policy
regimes. Under \emph{zero-image} the vision tower sees a constant
all-zero input and its output is approximately a constant bias, so
the projector head can learn to ignore the visual stream entirely
and recover whatever the language prompt, state and action priors
provide. Under \emph{shuffled-image} the vision tower sees
dynamically varying but spatially meaningless input, its output is
non-constant but \emph{misleading}, and the policy is forced to
integrate an incorrect visual signal rather than ignore it. We
therefore expect the shuffled-image MSE to be \emph{above} the
zero-image MSE on cells where the encoder is genuinely using
fine-grained spatial structure, and to be \emph{close to or below}
the zero-image MSE on cells where the encoder is mostly using
low-frequency image statistics (because zero-image is then the
heavier OOD shift).

The pattern we observe (\autoref{tab:sanity_controls}) is
backbone-and-suite-dependent and falls into three regimes.
\emph{(i)} On SmolVLA-\texttt{spatial} and SmolVLA-\texttt{object}
shuffled-image MSE ($0.10133$, $0.09216$) is substantially above
zero-image MSE ($0.07696$, $0.07638$), so the shuffle-gap
($+48.49\%$, $+51.16\%$) is more than double the vision-gap
($+12.77\%$, $+25.28\%$). The SmolVLA encoder
is therefore using real spatial-action correspondence on both
suites, not the pixel histogram --- forcing the policy to integrate
a histogram-matched but spatially scrambled image is strictly worse
than letting it ignore the visual stream.
\emph{(ii)} On $\pi_{0.5}$-\texttt{object} the shuffle-gap is
$+22.18\%$ while the vision-gap is $+95.54\%$, so destroying
spatial structure costs roughly a quarter of the full
visual-contribution budget. The SigLIP-vs-DINOv2 within-wrapper
cell-level encoder gap on this suite ($1.7$--$7\%$) sits well below
the $22\%$ shuffle-gap, so we read the
$\pi_{0.5}$-\texttt{object} encoder ranking as still validated by a
genuine vision-utility band: encoder differences on this suite are
inside the shuffle-tolerance band, but they are not within the
\emph{shuffle-gap} band itself.
\emph{(iii)} On $\pi_{0.5}$-\texttt{spatial} the shuffle-gap is
only $+3.90\%$ while the vision-gap is $+50.54\%$, so $\pi_{0.5}$
on \texttt{libero\_spatial} loses almost nothing when the spatial
structure is destroyed as long as the pixel histogram is preserved.
We make this limitation explicit:
the SigLIP-vs-DINOv2 cell-level gap on
$\pi_{0.5}$-\texttt{libero\_spatial} ($1.7$--$12\%$ MSE between
SigLIP and DINOv2 within the wrapper, depending on the cell) is
\emph{comparable to} the $3.9\%$ shuffle-tolerance band on the same
suite --- for several cells it is even smaller --- so the
within-wrapper SigLIP-vs-DINOv2 ranking on this
specific suite may reflect low-frequency image statistics (colour
distribution, texture marginals) rather than fine-grained
spatial-action correspondence. We therefore treat
$\pi_{0.5}$-\texttt{libero\_spatial} as the weakest of the supporting
cells in our main claim and limit the claim about it accordingly
in \autoref{sec:disc:limit}. The remaining three (backbone, suite)
cells (SmolVLA-\texttt{spatial}, SmolVLA-\texttt{object},
$\pi_{0.5}$-\texttt{object}) are validated by both the zero-image
and shuffled-image controls.

\paragraph{Vision usage is backbone-and-suite-dependent.}
Combining the zero-image and shuffled-image controls gives us a
sharper picture than either control alone: the way the grafted
encoder uses the visual input is not a single mechanism that
transfers across backbones and suites. SmolVLA on both suites uses
genuine spatial-action correspondence (large vision-gap, even
larger shuffle-gap). $\pi_{0.5}$ on \texttt{libero\_object} also
uses real spatial structure, but the shuffle-tolerance band
($+22\%$) is narrower than on SmolVLA and the encoder ranking is
correspondingly less robust to noise. $\pi_{0.5}$ on
\texttt{libero\_spatial} sits in a low-frequency-statistic-dominated
regime where the encoder ranking is shuffle-tolerant and where
within-wrapper encoder-comparison claims should be interpreted as
weaker than the SmolVLA cells. We treat this backbone-and-suite
dependence of vision usage as an unexpected secondary contribution
of this paper: vision-utility is not a property of the backbone or
of the suite in isolation but of the (backbone, suite) cell, and a
single shuffled-image control surfaces this dependence at low cost.
Within this caveat, the present zero-image-plus-shuffled-image
finding still rules out the degenerate case of ``vision is not used
at all'' on three of the four (backbone, suite) cells under the
fixed grafting protocol, which is the load-bearing claim we
make in this paper.

\subsection{Training-loss plateau evidence for the $2{,}000$-step budget}
\label{sec:exp:plateau}

\begin{figure}[t]
  \centering
  \includegraphics[width=0.95\columnwidth]{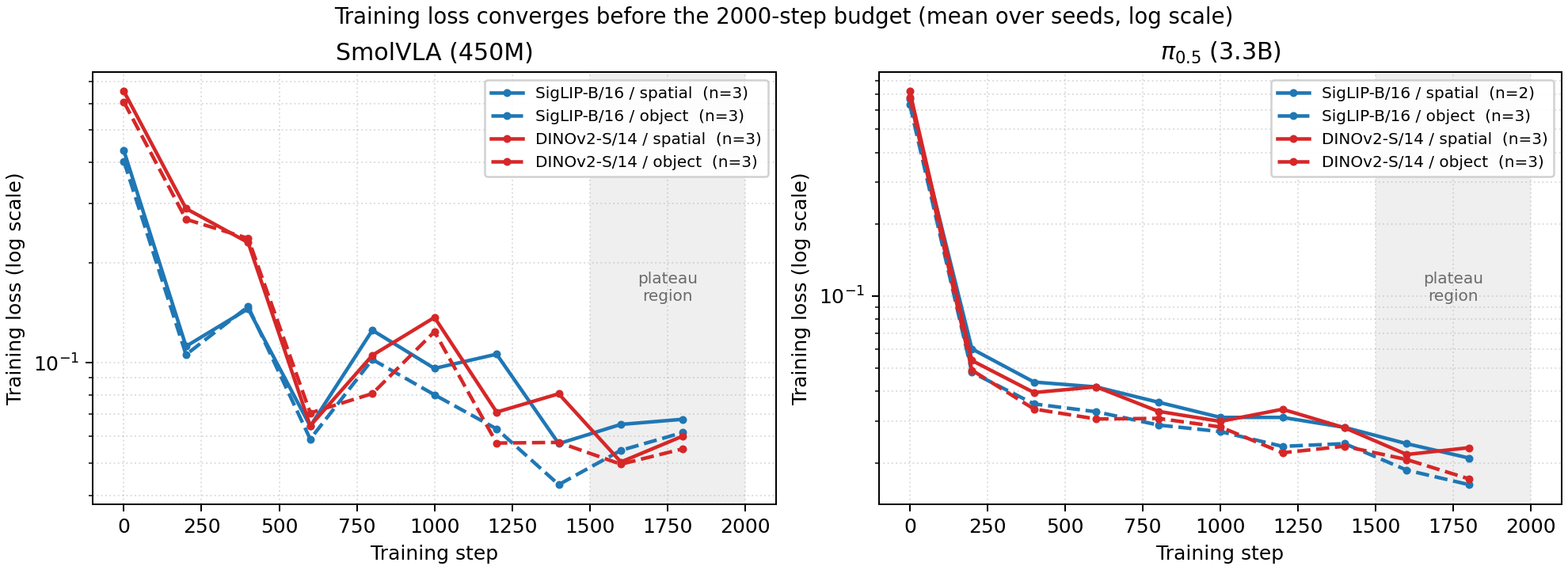}
  \caption{Training-loss curves for the four candidate encoders on
  SmolVLA and $\pi_{0.5}$ on the two LIBERO suites, seed $42$.
  All evaluated encoders reach a plateau region by step
  $1{,}500$--$2{,}000$ on both backbones, supporting our wording in
  \autoref{sec:method:train} that conclusions are reported under a
  fixed $2{,}000$-step projector budget rather than as ``best encoder
  absolutely''.}
  \label{fig:supp_loss}
\end{figure}

A standing question for any fixed-budget projector recipe is whether
the budget is the binding constraint on the encoder ranking. The
per-encoder training-loss curves for each (backbone, suite) pair in
\autoref{fig:supp_loss} confirm that all evaluated encoders
reach a plateau region by step $1{,}500$--$2{,}000$ on both backbones;
the $2{,}000$-step budget is not bottlenecking encoder ranking.

\section{Discussion}
\label{sec:discussion}

\subsection{Main takeaway}
\label{sec:disc:main}

Our central observation is descriptive and task-dependent: under a
controlled frozen-backbone grafting protocol, the small-backbone
encoder winner does not reliably select the large-backbone top tier.
SigLIP is the lowest-mean-MSE encoder on SmolVLA across both LIBERO
suites (with $3/3$ seeds in the SigLIP-vs-DINOv2 direction on each
suite). On $\pi_{0.5}$-\texttt{libero\_spatial} DINOv2-small is the
lowest-mean-MSE encoder (with $3/3$ seeds in the DINOv2-vs-SigLIP
direction, although the seed-44 cell is a numerical near-tie at
$\Delta = 0.00001$ MSE so the within-seed margin at that seed
should not be over-read). On $\pi_{0.5}$-\texttt{libero\_object} the
three strongest encoders (SigLIP $0.02149$, DINOv2 $0.02166$, FastViT
$0.02206$) sit in a near-tie band within $2.7\%$ relative of each
other, and the seed-level top-1 identity is not stable to a single
additional seed (SigLIP$<$DINOv2 at seed 44, DINOv2$<$SigLIP at seeds
42 and 43). The suite-level summary is that $3/4$ backbone-suite
SigLIP-vs-DINOv2 directions support the
backbone-dependent top-tier pattern, with
$\pi_{0.5}$-\texttt{libero\_object} as the lone exception; we treat
this $3/4$ as the load-bearing inferential summary, with the
$11/12$ seed-level cell split as descriptive support.
At the (backbone, suite) level, the sanity controls
(\autoref{sec:exp:sanity_controls}, \autoref{tab:sanity_controls})
validate three of the four cells under both the zero-image and
shuffled-image controls; the fourth,
$\pi_{0.5}$-\texttt{libero\_spatial}, is shuffle-tolerant
(shuffle-gap $+3.9\%$, comparable to the $1.7$--$12\%$ within-wrapper
SigLIP-vs-DINOv2 cell-level gap on the same suite), and we treat it as
the weakest supporting cell, with the implication discussed in
\autoref{sec:disc:limit} below. The finer
$11/12$ vs $1/12$ cell-level split is descriptive evidence consistent
with the same pattern (\autoref{tab:directional}); we do not defend it
as independent inferential evidence because the cells share seeds,
projector initialization, and validation episode order, and so they
are not pseudo-independent samples. We deliberately do \emph{not}
call this a full ranking inversion: the cross-backbone Spearman
correlation stays positive on both suites ($+0.80$, $+1.00$) because
the worst encoder in the pool is reliably the worst encoder on both
backbones. The claim we are willing to defend is narrower: the top-1
encoder selected on SmolVLA does not reliably select the top-1
encoder on $\pi_{0.5}$ under this offline diagnostic, and the
partial bottom-of-pool ordering transfers while the top-1 choice
does not. The $\pi_{0.5}$-\texttt{libero\_spatial} seed-44 near-tie
supports rather than weakens this claim, since it shows the
large-backbone top tier is a narrow-margin competition unsuited for
small-backbone top-1 commitment.
At a minimum this argues that future encoder-comparison papers should
report at least two backbones at different scales \emph{and} a
robustness check at a third seed before stating ``encoder $X$ is best
for VLA'' as an architectural prescription, and that a small grafting
sweep on the target released backbone is a cheap way to confirm or
refute the small-VLA top-1 choice before any expensive downstream
training.

A second observation emerges from the native-vs-grafted comparison in
\autoref{sec:exp:native_anchor} (\autoref{tab:native_anchor},
\autoref{fig:native_anchor}) once the sanity controls in
\autoref{sec:exp:sanity_controls} are folded in. On the two released
backbones we test, the sign of the native-vs-grafted offline MSE
comparison differs: on SmolVLA the best grafted encoder is
$30$--$32\%$ higher MSE than the native tower, while on $\pi_{0.5}$
the best grafted encoder is $42$--$43\%$ lower MSE than the native
PaliGemma SigLIP-So400m tower. The new sanity controls reveal that
the $\pi_{0.5}$ $42$--$43\%$ lower-MSE-than-native number is
\emph{wrapper-induced (interface-induced) rather than
candidate-encoder-substitution induced}, and therefore cannot be read
as encoder superiority: the
native-encoder-through-graft-interface control
(\autoref{tab:sanity_controls}) shows that
routing the native $\pi_{0.5}$ vision tower through the fixed
grafting protocol \emph{alone}, with no candidate encoder substitution,
already moves MSE from native by $-49.84\%$ on \texttt{spatial} and
$-51.94\%$ on \texttt{object} --- essentially recovering the
$42$--$43\%$ best-grafted gap without ever substituting an encoder.
We therefore reinterpret the $\pi_{0.5}$ native-vs-grafted gap as
``the wrapper itself is a non-neutral path that happens to help the
$\pi_{0.5}$ native tower'', not as ``a candidate encoder is
intrinsically better than the native tower''. Symmetrically, the
wrapper degrades the SmolVLA native path by $+45$--$56\%$, which
contributes to the $30$--$32\%$ best-grafted-vs-native gap on SmolVLA
in the opposite direction. The wrapper effect therefore has
opposite sign across the two backbones, and the candidate-encoder
comparison that survives this reinterpretation cleanly is the
within-wrapper SigLIP-vs-DINOv2 contrast, because both encoders
cross the same wrapper. We are deliberately careful not to
interpret the cross-backbone sign of the wrapper effect as a causal
scale effect either: SmolVLA and $\pi_{0.5}$ differ in many
architectural and pretraining factors besides parameter count, so
the most we can say descriptively is that the wrapper effect flips
sign between the two released checkpoints, and that the sign on
either checkpoint should not be assumed to transfer to a third
released backbone without re-measurement. The targeted pooling
ablation in \autoref{sec:exp:pool} further shows that the
SigLIP--DINOv2 ordering is preserved across three pooling choices on
SmolVLA-\texttt{libero\_spatial}-seed42, which rules out
token-pooling as an obvious explanation for the SmolVLA
SigLIP-over-DINOv2 result on that slice but does not establish
pooling invariance across all backbones, suites, encoders or seeds.

\subsection{Seed-44 sensitivity and the noise-dominated suite}
\label{sec:disc:seed44}

The seed-44 robustness check exposes a regime distinction between the
two LIBERO suites that we did not anticipate from the seed
$\{42, 43\}$ data alone, and that we think future encoder-comparison
papers should adopt as a routine sanity check. On
$\pi_{0.5}$-\texttt{libero\_spatial} the cross-encoder DINOv2-vs-SigLIP
gap is up to $13\%$ relative on the strongest seed in the
$\{42, 43, 44\}$ set and the direction is preserved across all three
seeds; we read this as a regime where the cross-encoder signal exceeds
seed noise. On $\pi_{0.5}$-\texttt{libero\_object} the three strongest
encoders are within $2.7\%$ relative of each other under the
three-seed mean ($0.02149$ vs $0.02166$ vs $0.02206$), the seed-level
top-1 identity flips between SigLIP and DINOv2 across the three
seeds, and a third seed at the SigLIP/DINOv2 cells alone is enough to
reverse the rank-order claim that the seed-$\{42, 43\}$ data alone
would have supported. This is itself part of our finding: the
cross-backbone top-1 question is meaningful only on suites where the
cross-encoder gap exceeds seed noise (\texttt{libero\_spatial}: gap
up to $13\%$, stable; $\pi_{0.5}$-\texttt{libero\_object}: gap
$< 3\%$ at the top of the pool, unstable). For the noise-dominated
suite the operationally honest statement is that the top three
encoders are within seed-level noise of each other, and the
appropriate selector is one of (a) running additional seeds until the
gap exceeds noise or (b) tabulating the near-tie band explicitly and
selecting on a secondary axis (latency, parameter count, deployment
constraints) rather than on offline MSE alone.

\subsection{Vision usage is backbone-and-suite-dependent}
\label{sec:disc:vision_usage}

A secondary observation that emerges only once the zero-image and
shuffled-image controls of \autoref{sec:exp:sanity_controls} are
both available is that the grafted encoder does not use the visual
input in a single uniform way across our four
(backbone, suite) cells. The four cells fall into three regimes,
which we treat as an unexpected contribution of this paper rather
than as a clean-up footnote on the main matrix.

\emph{Real spatial-action correspondence (SmolVLA, both suites).}
On SmolVLA the vision-gap is $+12.77$--$25.28\%$ and the shuffle-gap
is $+48.49$--$51.16\%$ --- substantially \emph{larger} than the
vision-gap. The encoder uses spatial structure that the
shuffled-image baseline destroys, and a histogram-matched but
spatially scrambled image is strictly worse than no image at all
because the projector head is forced to integrate a misleading
visual signal rather than learn to ignore it. The SmolVLA encoder
ranking on both suites is therefore validated as
spatial-correspondence-driven.

\emph{Real spatial-action correspondence with a narrower band
($\pi_{0.5}$-\texttt{object}).}
On $\pi_{0.5}$-\texttt{object} the vision-gap is $+95.54\%$ and the
shuffle-gap is $+22.18\%$, so destroying spatial structure costs
roughly a quarter of the visual-contribution budget. The
within-wrapper SigLIP-vs-DINOv2 cell-level encoder gap on this
suite ($1.7$--$7\%$) is below the $22\%$ shuffle-tolerance band,
so we read the $\pi_{0.5}$-\texttt{object} encoder ranking as
genuinely validated by a real-vision-utility band even if the band
is narrower than on SmolVLA.

\emph{Low-frequency image statistics dominated
($\pi_{0.5}$-\texttt{spatial}).}
On $\pi_{0.5}$-\texttt{spatial} the vision-gap is $+50.54\%$ but
the shuffle-gap is only $+3.90\%$. The encoder loses almost nothing
when the spatial structure is destroyed as long as the pixel
histogram is preserved, which is direct evidence that the encoder
is using low-frequency image statistics (colour distribution,
texture marginals) rather than fine-grained spatial-action
correspondence on this specific suite. The within-wrapper
SigLIP-vs-DINOv2 cell-level gap on this suite ($1.7$--$12\%$) is
\emph{larger} than the shuffle-tolerance band, so the SigLIP-vs-DINOv2
ranking on this cell may reflect which encoder happens to emit a
more useful low-frequency activation distribution rather than which
encoder reads spatial structure better. We
flag this in \autoref{sec:disc:limit} below.

The operational reading is that vision-utility is a property of
the (backbone, suite) cell rather than of the backbone alone or
the suite alone. A successor study evaluating encoders on a single
(backbone, suite) cell could therefore draw qualitatively
different conclusions depending on which cell it picks, and the
shuffled-image control turns out to be a cheap but high-information
test that surfaces this dependence at the cost of a single
additional projector run per cell. We recommend that future
encoder-comparison studies run both the zero-image and
shuffled-image controls as a routine pair.

\subsection{Hypotheses for the wrapper non-neutrality and the
            cross-backbone sign difference}
\label{sec:disc:why}

The native-encoder-through-graft-interface control of
\autoref{sec:exp:sanity_controls} shows
that the cross-backbone sign of the native-vs-grafted MSE difference
is substantially attributable to the fixed grafting protocol
itself being non-neutral
with opposite sign across the two backbones (wrapper-gap
$+45.45\%$ to $+55.81\%$ on SmolVLA, $-49.84\%$ to $-51.94\%$ on
$\pi_{0.5}$).
We do not have a mechanistic explanation for this opposite-sign
wrapper effect, but we offer two hypotheses that are consistent with
the sanity-control data and testable in future work. First, the
SmolVLM SigLIP vision tower already emits a $(B, 64, 960)$ token grid
matched to the SmolVLA expert, so feeding it through an additional
$8 \times 8$ AdaptiveAvgPool and a fresh linear projector strictly
removes information --- there is no compression bottleneck for the
wrapper to add value at on this backbone. By contrast, the
PaliGemma SigLIP-So400m tower on $\pi_{0.5}$ emits a higher-dimensional
$(B, 256, 2048)$ grid, and the wrapper's $16 \times 16$ pool and
linear projector appear to provide a useful bottleneck or
regularisation rather than a destructive compression. Second, on
$\pi_{0.5}$ the Gemma expert attends across $256$ tokens and may be
more sample-efficient at re-interpreting an unfamiliar token stream
than the SmolVLM expert is at $64$ tokens, so the marginal cost of
crossing an extra projector may shrink (and even flip to a benefit)
once the expert is large enough to compensate. Both hypotheses
predict that the wrapper-gap should shrink as the projector capacity
grows and as the expert hidden size grows, and we leave the
quantitative tests to future work.

\subsection{Limitations}
\label{sec:disc:limit}

This study has several limitations that we wish to be explicit about
rather than hide.

\paragraph{Preprocessing-protocol confound.}
All grafted runs in this paper share a unified $224{\times}224$
preprocessing path (bilinear resize, the unified normalization
inherited from the backbone's vision-input pipeline) so that every
encoder is exposed to the same input distribution. This unified path
may understate encoders whose pretrained normalization or native
input resolution differs from the unified protocol --- in particular,
DINOv2 is natively trained at $518{\times}518$ with ImageNet
normalization, whereas SigLIP base/16 is natively at $224{\times}224$
with $\mu{=}\sigma{=}0.5$ normalization. Preprocessing sensitivity is
a known threat to validity that we do not quantify here; the present
ranking should be read as conditional on the unified preprocessing
path described in \autoref{sec:method:graft}, and a per-encoder
official-preprocessing comparison is a natural follow-up.

\paragraph{No closed-loop success claim under embodiment mismatch.}
We do not report closed-loop success because native released
checkpoints collapse under embodiment mismatch; therefore this paper
makes no deployment or success-rate claim. The action-MSE numbers in
this paper are an offline diagnostic and should not be used as
evidence of policy deployment readiness.

\paragraph{Shuffle-tolerance of the $\pi_{0.5}$-\texttt{spatial} ranking.}
We note that the within-wrapper SigLIP-vs-DINOv2 encoder ranking on
$\pi_{0.5}$-\texttt{libero\_spatial} should be interpreted
cautiously: the shuffled-image MSE on this cell is only $3.9\%$
worse than the real-image MSE, comparable to or smaller than the
$1.7$--$12\%$ within-wrapper SigLIP-vs-DINOv2 cell-level gap on the
same suite. This means the SigLIP-vs-DINOv2 ranking on
$\pi_{0.5}$-\texttt{libero\_spatial} may reflect low-frequency
image statistics rather than fine-grained spatial-action
correspondence, and we treat this cell as the weakest supporting
cell in our main claim. The other three
cells (SmolVLA-\texttt{spatial}, SmolVLA-\texttt{object},
$\pi_{0.5}$-\texttt{object}) remain validated by both the
zero-image \emph{and} shuffled-image controls of
\autoref{sec:exp:sanity_controls}: the $\pi_{0.5}$-\texttt{object}
vision usage is validated by the larger $22\%$
shuffle-tolerance band on that cell. The shuffled-image control is
precisely the test that surfaces this fragility on
$\pi_{0.5}$-\texttt{spatial}.

\paragraph{Two-to-three seeds per cell.}
Each (backbone, suite, encoder) cell has $N \in \{2, 3\}$ seeds:
SigLIP and DINOv2 are at $n{=}3$ (seeds 42, 43, 44), and FastViT and
RepViT are at $n{=}2$ (seeds 42, 43, with seed 44 deferred to future
work). With few seeds, standardized effect sizes are unstable and
paired $t$-tests have low power; we therefore report directional
consistency as our load-bearing statistic and treat Cohen's $d$ and
paired $t$-values as descriptive only. On
$\pi_{0.5}$-\texttt{libero\_object} the top three encoders (SigLIP
$0.02149$, DINOv2 $0.02166$, FastViT $0.02206$) form a near-tie band
where the seed-level top-1 identity flips across the three seeds; a
fourth seed on the SigLIP-vs-DINOv2 pair on $\pi_{0.5}$-\texttt{object}
is an important follow-up.

\paragraph{Offline action MSE rather than closed-loop success.}
We report validation action MSE on a held-out episode split rather
than closed-loop success rates. The reason is that the released
SmolVLA and $\pi_{0.5}$ checkpoints were trained on SO-100 trajectories
with a six-degree-of-freedom action schema, while LIBERO uses a
seven-degree-of-freedom Franka schema; zero-shot closed-loop success
is therefore near zero for both native and grafted variants and would
provide essentially no signal. Offline action MSE is the right proxy
under this mismatch --- it measures whether the grafted encoder lets
the backbone reproduce the demonstration policy --- but it is not a
substitute for embodied success, and offline MSE does not necessarily
correlate with closed-loop rollout success because of compounding
error, state-distribution shift and multi-modal action ambiguity. We
therefore frame every conclusion in this paper as an
\emph{offline diagnostic} under embodiment mismatch and do not claim
closed-loop deployment improvement.

\paragraph{Two released backbones at different scales, not isolated scale.}
We test two released VLA backbones (SmolVLA-450M and $\pi_{0.5}$-3.3B)
that differ in many factors besides parameter count: language
backbone, action expert, hidden size, token count, native vision
tower, pretraining data, projector architecture and native image
preprocessing. We therefore describe the native-vs-grafted sign
difference as occurring \emph{across the two released backbones we
test}, not as caused by backbone scale; the comparison is
scale-correlated but architecture-confounded.

\paragraph{Projector bottleneck risk.}
The grafted policy has only a single linear projector
($0.37$--$1.58$M parameters depending on encoder output dimension and
target backbone hidden size). One might worry that the ranking we
observe is really a ranking of which encoder happens to be most
``linear-projectable'' into the backbone's hidden space, rather than a
ranking of encoder quality. The gate-level LoRA ablation in
\autoref{sec:exp:p2_lora} improves all four tested cells, which is
qualitative evidence against a projector capacity ceiling driving the
ranking, but the ablation has only one seed, one suite, and two of
four encoders, so it does not control the cross-encoder ranking under
P2.

\paragraph{Frozen-LM path dependence.}
Our main matrix freezes the language model and the action expert.
This is required for a fair encoder comparison but it means our
top-1 observation is strictly under \emph{zero gradient flow into the
backbone}. A LoRA fine-tune of the expert might allow a weaker encoder
to ``catch up'' or it might further amplify the gap; we cannot tell
from the current data.

\paragraph{Wrapper non-neutrality as a methodological contribution.}
The native-encoder-through-graft-interface control of
\autoref{sec:exp:sanity_controls}
demonstrates that the fixed grafting protocol itself shifts
native MSE by
$+45$--$56\%$ on SmolVLA and by $-50$--$52\%$ on $\pi_{0.5}$ at
seed $42$, with opposite sign across the two backbones. The
within-interface SigLIP-vs-DINOv2 ranking in
\autoref{sec:exp:main_table} therefore remains the cleanest
experimental contrast in this paper because both encoders cross the
same fixed grafting protocol, but the native-vs-grafted comparison in
\autoref{sec:exp:native_anchor} can no longer be read as
``a candidate encoder is intrinsically better than the native vision
tower'' on $\pi_{0.5}$ --- a substantial portion of the
$42$--$43\%$ best-grafted-minus-native gap is explained by the
interface alone before any candidate encoder substitution. We retain
the native-vs-grafted table as a descriptive contrast and rely on
the native-through-interface and zero-image controls for the
load-bearing sanity-control statements. The wrapper
non-neutrality is itself a methodological contribution:
benchmarks that swap visual
encoders without controlling for the swap interface conflate encoder
choice with interface-backbone compatibility, and reporting the
fixed-grafting-protocol caveat (plus the
native-through-interface control number) is now part of the
diagnostic protocol we recommend for any successor encoder-comparison
study on released VLA backbones.

\paragraph{P2 LoRA matrix is gate-level.}
The P2 ablation reported in \autoref{sec:exp:p2_lora} is deliberately
gate-level: one seed, one suite (\texttt{libero\_spatial}), two of the
four encoders, two backbones ($2$ encoders $\times$ $2$ backbones
$=4$ cells in total, no P2 DINOv2 and no P2 RepViT). The seed-$42$ P2 SigLIP cell on $\pi_{0.5}$
comes in below the seed-matched P1 DINOv2-small cell, which is
suggestive of a P1$\to$P2 ordering shift, but the comparison crosses
two protocols at a single seed and is not a controlled cross-encoder
ranking test under P2. A controlled statement requires a full P2
sweep that matches the main matrix design ($4$ encoders $\times$ $2$
suites $\times$ $\geq 2$ seeds per backbone), which we leave to
future work.

\subsection{Future work}
\label{sec:disc:future}

Three directions are most pressing. First, extending the
seed-44 robustness check to the FastViT and RepViT cells; combined
with the existing seed-44 SigLIP/DINOv2 data this would give a fully
uniform $n{=}3$ matrix across the $4$ encoders and upgrade those two
cells from exploratory to confirmatory. Second, a follow-up
on the shuffle-tolerant $\pi_{0.5}$-\texttt{libero\_spatial} cell
identified by the shuffled-image control of
\autoref{sec:exp:sanity_controls} with a finer-grained baseline
sweep (per-channel histogram-only, low-pass-only, and patch-level
shuffle variants) to pin down which low-frequency statistic is
driving the encoder ranking on that suite. Third, once an
embodiment-matched
dataset is available, a true closed-loop evaluation
with the same fixed grafting protocol to test whether the
offline top-1 observation translates to closed-loop success in the same
direction; this is the only experiment that can upgrade the current
claim from an offline diagnostic to a deployment-relevant prescription.

{\scriptsize
\renewcommand{\baselinestretch}{0.92}\selectfont
\bibliographystyle{abbrv}
\bibliography{refs}

\begin{thebibliography}{10}

\bibitem{ba2016layer}
J.~L. Ba, J.~R. Kiros, and G.~E. Hinton.
\newblock Layer normalization.
\newblock {\em arXiv preprint arXiv:1607.06450}, 2016.

\bibitem{paligemma2024}
L.~Beyer, A.~Steiner, A.~S. Pinto, A.~Kolesnikov, X.~Wang, D.~Salz, M.~Neumann, I.~Alabdulmohsin, M.~Tschannen, E.~Bugliarello, et~al.
\newblock {PaliGemma}: A versatile 3{B} {VLM} for transfer.
\newblock {\em arXiv preprint arXiv:2407.07726}, 2024.

\bibitem{vla02024}
J.~Bjorck, F.~Castaneda, N.~Cherniadev, X.~Da, R.~Ding, L.~Fan, Y.~Fang, D.~Fox, F.~Hu, S.~Huang, et~al.
\newblock {VLA-0}: Building state-of-the-art {VLA}s with zero modification.
\newblock {\em arXiv preprint arXiv:2510.13054}, 2025.

\bibitem{pi05_2024}
P.~Intelligence, K.~Black, N.~Brown, D.~Driess, A.~Esmail, M.~Equi, C.~Finn, N.~Fusai, M.~Y. Galliker, D.~Ghosh, et~al.
\newblock $\pi_{0.5}$: a vision-language-action model with open-world generalization.
\newblock {\em arXiv preprint arXiv:2504.16054}, 2025.

\bibitem{openvla_oft2024}
M.~J. Kim, C.~Finn, and P.~Liang.
\newblock Fine-tuning vision-language-action models: Optimizing speed and success.
\newblock {\em arXiv preprint arXiv:2502.19645}, 2025.

\bibitem{openvla2024}
M.~J. Kim, K.~Pertsch, S.~Karamcheti, T.~Xiao, A.~Balakrishna, S.~Nair, R.~Rafailov, E.~Foster, G.~Lam, P.~Sanketi, et~al.
\newblock {OpenVLA}: An open-source vision-language-action model.
\newblock {\em arXiv preprint arXiv:2406.09246}, 2024.

\bibitem{liberopro2024}
S.~Li, S.~Zhao, J.~Wu, Y.~Lyu, B.~Liu, and Y.~Zhu.
\newblock {LIBERO}-pro: Benchmarking generalization for vision-language-action models.
\newblock {\em arXiv preprint arXiv:2509.21788}, 2025.

\bibitem{liberobench2023}
B.~Liu, Y.~Zhu, C.~Gao, Y.~Feng, Q.~Liu, Y.~Zhu, and P.~Stone.
\newblock {LIBERO}: Benchmarking knowledge transfer for lifelong robot learning.
\newblock {\em Advances in Neural Information Processing Systems}, 2023.

\bibitem{loshchilov2017adamw}
I.~Loshchilov and F.~Hutter.
\newblock Decoupled weight decay regularization.
\newblock {\em arXiv preprint arXiv:1711.05101}, 2017.

\bibitem{smolvlm2024}
A.~Marafioti, O.~Zohar, M.~Farr{\'e}, M.~Noyan, E.~Bakouch, P.~Cuenca, C.~Zakka, L.~B. Allal, A.~Lozhkov, N.~Tazi, et~al.
\newblock {SmolVLM}: Redefining small and efficient multimodal models.
\newblock {\em arXiv preprint arXiv:2504.05299}, 2025.

\bibitem{oquab2023dinov2}
M.~Oquab, T.~Darcet, T.~Moutakanni, H.~Vo, M.~Szafraniec, V.~Khalidov, P.~Fernandez, D.~Haziza, F.~Massa, A.~El-Nouby, et~al.
\newblock {DINOv2}: Learning robust visual features without supervision.
\newblock {\em arXiv preprint arXiv:2304.07193}, 2023.

\bibitem{theia2024}
J.~Shang, K.~Schmeckpeper, B.~B. May, M.~V. Minniti, T.~Kelestemur, D.~Watkins, and L.~Herlant.
\newblock {Theia}: Distilling diverse vision foundation models for robot learning.
\newblock {\em arXiv preprint arXiv:2407.20179}, 2024.

\bibitem{smolvla2024}
M.~Shukor, D.~Aubakirova, F.~Capuano, P.~Kooijmans, S.~Palma, A.~Zouitine, M.~Aractingi, C.~Pascal, M.~Russi, A.~Marafioti, et~al.
\newblock {SmolVLA}: A vision-language-action model for affordable and efficient robotics.
\newblock {\em arXiv preprint arXiv:2506.01844}, 2025.

\bibitem{vasu2023fastvit}
P.~K.~A. Vasu, J.~Gabriel, J.~Zhu, O.~Tuzel, and A.~Ranjan.
\newblock {FastViT}: A fast hybrid vision transformer using structural reparameterization.
\newblock {\em arXiv preprint arXiv:2303.14189}, 2023.

\bibitem{wang2023repvit}
A.~Wang, H.~Chen, Z.~Lin, J.~Han, and G.~Ding.
\newblock {RepViT}: Revisiting mobile {CNN} from {ViT} perspective.
\newblock {\em arXiv preprint arXiv:2307.09283}, 2023.

\bibitem{vlm4vla2024}
T.-H. Wang, A.~Maalouf, G.~Rosman, S.~Karaman, and D.~Rus.
\newblock On the choice of vision-language backbone for robot policy learning.
\newblock {\em arXiv preprint arXiv:2405.05956}, 2024.

\bibitem{liberoplus2024}
S.~Wu, S.~Liu, Y.~Lyu, S.~Liu, Y.~Wu, J.~Sun, C.~Cui, P.~P. Liang, Y.-H. Wu, and J.~Zhou.
\newblock {LIBERO}-plus: In-depth robustness analysis of vision-language-action models.
\newblock {\em arXiv preprint arXiv:2511.17850}, 2025.

\bibitem{siglip2023}
X.~Zhai, B.~Mustafa, A.~Kolesnikov, and L.~Beyer.
\newblock Sigmoid loss for language image pre-training.
\newblock {\em arXiv preprint arXiv:2303.15343}, 2023.

\end{thebibliography}
\par}

\end{document}